\documentclass[11pt]{article}

\usepackage[T1]{fontenc}  %
\usepackage{lmodern}      %
\usepackage[margin=1in]{geometry}
\usepackage{graphicx}
\usepackage{booktabs}
\usepackage{hyperref}
\usepackage{xcolor}
\usepackage{microtype}
\usepackage{amsmath,amssymb}
\usepackage{enumitem}
\usepackage{authblk}
\usepackage{fontawesome5}
\usepackage{tikz}
\usetikzlibrary{positioning,shapes.geometric,arrows.meta,calc,backgrounds,fit}
\usepackage{xspace}
\usepackage[numbers]{natbib}  %
\usepackage[section]{placeins}  %

\DeclareRobustCommand{\rev}[1]{#1}
\DeclareRobustCommand{\auditrev}[1]{#1}
\newenvironment{revision}{}{}
\hypersetup{hidelinks}

\newcommand{\projectname}{HealthAgentBench\xspace}

\newcommand{\taskXray}{X-ray Report Correction\xspace}
\newcommand{\taskTumor}{Pathology Tumor Area Selection\xspace}
\newcommand{\taskCt}{CT Abnormality Classification\xspace}

\newcommand{\taskDq}{EHR Data Quality Auditing\xspace}
\newcommand{\taskEhrshot}{EHR Event Modelling\xspace}

\newcommand{\taskMri}{\rev{MRI QA}\xspace}
\newcommand{\taskSlide}{Pathology Tumor Slide Selection\xspace}
\newcommand{\taskCqa}{EHR Compositional QA\xspace}
\newcommand{\taskSurvival}{Cancer Survival Prediction\xspace}

\def\countXray{10}
\def\countCt{10}
\def\countMri{10}
\def\countTumor{10}
\def\countSlide{10}
\def\countCqa{6}
\def\countDq{8}
\def\countEhrshot{6}
\def\countTte{5}
\def\countTotal{75}

\newcommand{\numtasks}{\rev{\countTotal}\xspace} %
\newcommand{\numtaskcategories}{\rev{9}\xspace} %

\newcommand{\bestResolution}{\rev{57.3\%}\xspace}

\title{\projectname: A Unified Benchmark Suite of Realistic Agentic Healthcare Environments for Challenging Frontier AI Agents}
\author{%
\makebox[\textwidth][c]{%
\begin{minipage}{1.22\textwidth}\centering
\mbox{Qianchu Liu}\thanks{Equal contribution.}~,
\mbox{Sheng Zhang\textsuperscript{*}},
\mbox{Guanghui Qin\textsuperscript{*}},\\
\mbox{Jeya Maria Jose Valanarasu},
\mbox{Xiaoke Huang},
\mbox{Peniel Argaw},
\mbox{Maximilian Rokuss},\\
\mbox{Mingyu Lu},
\mbox{Timothy Ossowski},
\mbox{Juan Manuel Zambrano Chaves},\\
\mbox{Cliff Wong},
\mbox{Naisargi Patel},
\mbox{Yashna Hasija},
\mbox{Mu Wei},
\mbox{Wen-wai Yim},\\
\mbox{Qin Liu},
\mbox{Zilin Jing},
\mbox{Jason Entenmann},
\mbox{Naoto Usuyama},
\mbox{Tristan Naumann},
\mbox{Hoifung Poon}
\end{minipage}}%
}
\affil{Microsoft Research}
\date{\href{https://microsoft.github.io/HealthAgentBench/}{microsoft.github.io/HealthAgentBench/}}

\begin{document}
\renewcommand{\thefootnote}{\fnsymbol{footnote}}
\maketitle
\renewcommand{\thefootnote}{\arabic{footnote}}
\setcounter{footnote}{0}

\begin{abstract}
As AI agents become increasingly capable of complex, long-horizon reasoning, rigorous evaluation is essential for measuring progress toward real-world healthcare applications. However, existing healthcare benchmarks are often static or narrow in clinical scope, \rev{motivating broader evaluation of agents acting on raw patient data.} We introduce \textbf{\projectname}, a suite of \numtasks agentic healthcare tasks across \numtaskcategories categories each with its unique environment. The benchmark suite spans \rev{\auditrev{diagnosis, specimen assessment, research, and data management workflows}} and a broad range of modalities. Each task is designed to replicate an end-to-end clinical workflow: given \rev{task instructions and supporting resources}, an agent must explore raw healthcare data, operate within a complex environment, and execute multi-step solutions that go beyond naive prompting, such as querying large clinical databases or interpreting gigapixel pathology images. \rev{All tasks are scored using binary success criteria derived from reference labels, executable rules, or computational baselines. We evaluate six agents (three GPT models through Codex and three Claude models through Claude Code) with one scored attempt per task (450 evaluations). Overall success ranges from 32.0\% to 57.3\%; Claude Code Opus-5.5 leads with a 57.3\% success rate, followed by GPT-6-astra and Sonnet-5.5 at 50.7\% each. Performance varies substantially by category with specialized imaging modalities such as CT and MRI being \auditrev{especially} difficult. HealthAgentBench provides a challenging and realistic benchmark with substantial room for future progress. We release our benchmark at}\url{https://github.com/microsoft/HealthAgentBench}.

\end{abstract}

\newpage
\section{Introduction}

\begin{figure}[t]
  \centering
  \scalebox{0.96}{%
  \begin{tikzpicture}[
    font=\sffamily\small,
    docker/.style={
      draw=blue!45!black, line width=1.2pt, rounded corners=6pt,
      fill=blue!2, inner sep=10pt,
    },
    docker title/.style={
      font=\sffamily\bfseries\small,
      text=blue!45!black,
    },
    stage/.style={
      draw=orange!70!black, line width=0.7pt, rounded corners=4pt,
      fill=orange!22, inner sep=6pt, align=center,
      minimum width=8.4cm,
    },
    stagetitle/.style={
      font=\sffamily\bfseries\footnotesize,
      text=white, fill=orange!75!black,
      rounded corners=2pt, inner xsep=5pt, inner ysep=2pt,
    },
    inset/.style={
      draw=orange!55!black, line width=0.4pt, fill=white,
      rounded corners=2pt, inner sep=5pt, align=left,
    },
    termbox/.style={
      draw=black!80, fill=black!90, text=green!75!black,
      font=\ttfamily\scriptsize, align=left,
      rounded corners=3pt, inner sep=6pt,
    },
    termtitle/.style={
      draw=black!80, fill=black!75, text=white,
      font=\ttfamily\tiny\bfseries, align=left,
      rounded corners=2pt, inner xsep=5pt, inner ysep=2pt,
    },
    evalbox/.style={
      draw=blue!45!black, line width=1pt, rounded corners=5pt,
      fill=blue!2, inner sep=8pt,
    },
    evalcard/.style={
      draw=orange!70!black, line width=0.7pt, rounded corners=3pt,
      fill=orange!22, inner sep=5pt, align=center,
      font=\sffamily\footnotesize\bfseries,
    },
    modality/.style={
      draw=black!55, rounded corners=2.5pt, thick,
      text width=3.0cm, minimum width=3.4cm, minimum height=1.32cm,
      align=center, fill=white, inner sep=3pt,
      font=\sffamily\scriptsize,
      execute at begin node={\hyphenpenalty=10000\exhyphenpenalty=10000},
    },
    pipe/.style={densely dotted, thick, blue!45!black, opacity=0.55},
  ]

  \node[stage] (stage1) at (0, 0) {%
    \vspace{1pt}\\
    \begin{minipage}{8cm}\centering
      \vspace{1pt}
      \tikz\node[inset, minimum width=7.6cm]{%
        \begin{minipage}{7.2cm}\ttfamily\scriptsize
          /data/patient/\\
          \hspace*{0.6em}|--- study\_01\_2187-10-28\_14-51-03/\\
          \hspace*{1.6em}|--- view\_01.jpg\\
          \hspace*{1.6em}|--- view\_02.jpg\\
          \hspace*{1.6em}`--- report.txt~~\textit{(full prior report)}\\
          \hspace*{0.6em}|--- study\_02\_2187-11-11\_17-18-35/\\
          \hspace*{1.6em}`--- ...\\
          \hspace*{0.6em}`--- \underline{target\_study}/\\
          \hspace*{1.6em}|--- view\_01.jpg\\
          \hspace*{1.6em}`--- report.txt~~\textit{(corrupted draft FINDINGS)}
        \end{minipage}%
      };
    \end{minipage}
  };
  \node[stagetitle, anchor=south, yshift=2pt] (stage1title)
    at (stage1.north) {Staging patient data and artefacts};

  \node[stage, below=1.50cm of stage1] (stage2) {%
    \begin{minipage}{8cm}\centering
      \tikz\node[inset, minimum width=7.6cm]{%
        \begin{minipage}{7.2cm}\sffamily\footnotesize\itshape
          A junior radiologist's draft FINDINGS for the target study
          may contain clinical errors. Review and correct it using the
          chest X-ray images and prior studies, then write the corrected
          report to \texttt{/workspace/submission.json}.
        \end{minipage}%
      };
    \end{minipage}
  };
  \node[stagetitle, anchor=south, yshift=2pt] (stage2title)
    at (stage2.north) {Provide minimal instruction};

  \node[stage, below=1.50cm of stage2] (stage3) {%
    \begin{minipage}{8cm}\centering
      \tikz\node[termbox, minimum width=7.6cm]{%
        \begin{minipage}{7.0cm}\ttfamily\scriptsize
          \textcolor{green!75!black}{agent\$ view target\_study/view\_01.jpg}\\
          \textcolor{white!80!black}{> 1024x1024 PA frontal, lung fields clear ...}\\
          \textcolor{green!75!black}{agent\$ zoom view\_01.jpg 4x --crop}\\
          \textcolor{white!80!black}{> left base: no consolidation; draft wrong}\\
          \textcolor{green!75!black}{agent\$ read target\_study/report.txt}\\
          \textcolor{white!80!black}{> DRAFT FINDINGS: left basal consolidation ...}\\
          \textcolor{green!75!black}{agent\$ read study\_01\_.../report.txt}\\
          \textcolor{white!80!black}{> prior: chronic left basal scarring, stable}\\
          \textcolor{green!75!black}{agent\$ write /workspace/submission.json}\\
          \textcolor{white!80!black}{> corrected FINDINGS submitted [ok]}
        \end{minipage}%
      };
    \end{minipage}
  };
  \node[stagetitle, anchor=south, yshift=2pt] (stage3title)
    at (stage3.north) {Agent execution};

  \node[fit=(stage1)(stage1title)(stage2)(stage2title)(stage3)(stage3title),
        inner xsep=10pt, inner ysep=10pt, outer sep=0pt] (dockercontent) {};

  \node[modality, anchor=north west] (e1)
    at ([xshift=1.2cm, yshift=0cm]dockercontent.north east)
    {\includegraphics[width=1.7cm,height=0.50cm,keepaspectratio]{assets/xray.png}\\[1pt]%
     {\bfseries \taskXray}\\[0.5pt]{\tiny X-ray + report \textbullet\ \countXray\ tasks}};
  \node[modality, below=0.05cm of e1] (e2)
    {\includegraphics[width=1.7cm,height=0.50cm,keepaspectratio]{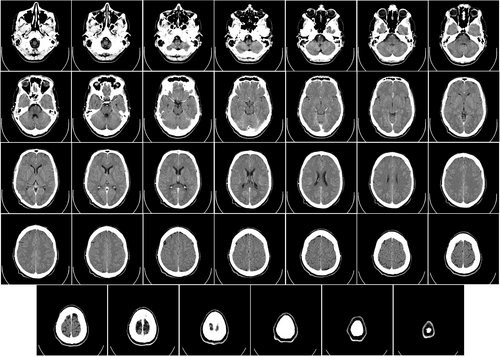}\\[1pt]%
     {\bfseries \taskCt}\\[0.5pt]{\tiny 3D CT volume \textbullet\ \countCt\ tasks}};
  \node[modality, below=0.05cm of e2] (e3)
    {\includegraphics[width=1.7cm,height=0.50cm,keepaspectratio]{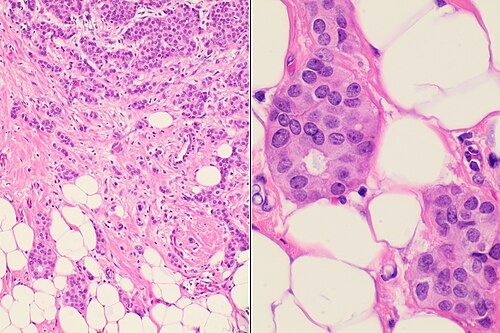}\\[1pt]%
     {\bfseries \taskTumor}\\[0.5pt]{\tiny Whole-slide image \textbullet\ \countTumor\ tasks}};
  \node[modality, below=0.05cm of e3] (e4)
    {\includegraphics[width=1.7cm,height=0.50cm,keepaspectratio]{assets/pathology.jpg}\\[1pt]%
     {\bfseries \rev{\taskSlide}}\\[0.5pt]{\tiny \rev{Eight whole slides \textbullet\ \countSlide\ tasks}}};
  \node[modality, below=0.05cm of e4] (e5)
    {\includegraphics[width=1.7cm,height=0.50cm,keepaspectratio]{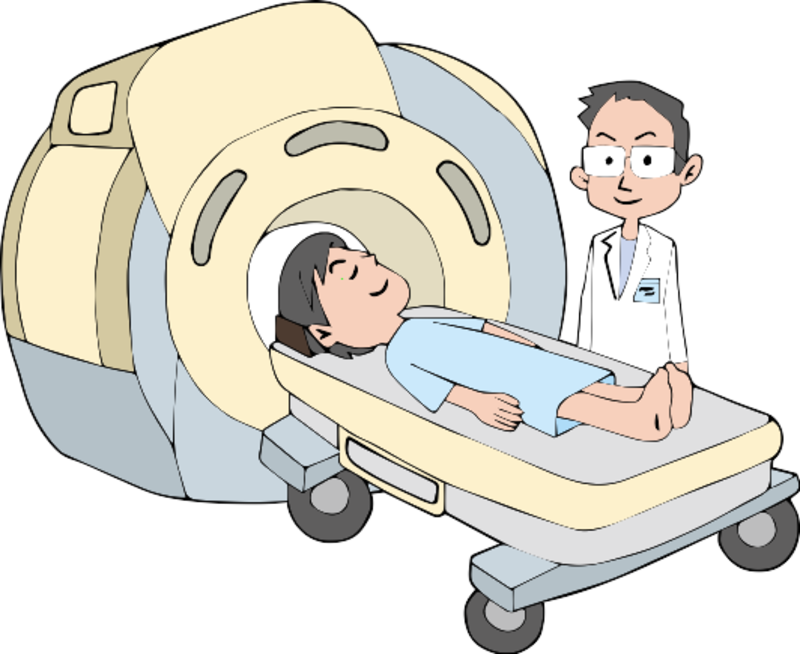}\\[1pt]%
     {\bfseries \rev{\taskMri}}\\[0.5pt]{\tiny \rev{3D MRI arrays \textbullet\ \countMri\ tasks}}};
  \node[modality, below=0.05cm of e5] (e6)
    {{\large\faTable}\\[1pt]%
     {\bfseries \taskDq}\\[0.5pt]{\tiny Tabular EHR \textbullet\ \countDq\ tasks}};
  \node[modality, below=0.05cm of e6] (e7)
    {{\large\faChartLine}\\[1pt]%
     {\bfseries \taskEhrshot}\\[0.5pt]{\tiny Longitudinal EHR \textbullet\ \countEhrshot\ tasks}};
  \node[modality, below=0.05cm of e7] (e8)
    {{\large\faDatabase}\\[1pt]%
     {\bfseries \rev{\taskCqa}}\\[0.5pt]{\tiny \rev{EHR + questions \textbullet\ \countCqa\ tasks}}};
  \node[modality, below=0.05cm of e8] (e9)
    {{\large\faChartLine}\\[1pt]%
     {\bfseries \rev{\taskSurvival}}\\[0.5pt]{\tiny \rev{Clinical tables \textbullet\ \countTte\ tasks}}};

  \coordinate (dockernw) at (dockercontent.north west |- e1.north);
  \coordinate (dockerse) at (dockercontent.south east |- e9.south);
  \begin{scope}[on background layer]
    \node[docker, fit=(dockernw)(dockerse), inner sep=0pt] (docker) {};
  \end{scope}
  \node[docker title, fill=blue!45!black, text=white,
        rounded corners=2pt, inner xsep=6pt, inner ysep=2pt,
        anchor=south, yshift=3pt]
    at (docker.north) {Task environment (Docker container)};

  \foreach \m in {e1, e2, e3, e4, e5, e6, e7, e8, e9} {
    \draw[pipe] (\m.west) -- ($(\m.west)+(-1.1cm,0)$);
  }

  \node[rotate=-90, anchor=center, align=center,
        font=\sffamily\bfseries\footnotesize, text=white,
        fill=blue!45!black, rounded corners=2pt, inner xsep=7pt, inner ysep=3pt]
    at ([xshift=0.5cm]$(e1.east)!0.5!(e9.east)$)
    {\numtasks tasks across \numtaskcategories task categories with unique environments};

  \node[evalcard, anchor=north west,
        minimum width=4.6cm, minimum height=0.55cm,
        inner xsep=6pt, inner ysep=3pt]
    at ([yshift=-0.9cm, xshift=0pt]docker.south west)
    (evalcard1) {\rev{Hidden reference labels}};
  \node[evalcard, right=0.25cm of evalcard1,
        minimum width=4.8cm, minimum height=0.55cm,
        inner xsep=6pt, inner ysep=3pt]
    (evalcard2) {Evaluation script (post-process \& score)};

  \node[anchor=west, font=\sffamily\footnotesize\bfseries, text=blue!45!black]
    at ([xshift=0.45cm]evalcard2.east) (outlbl) {Output:};
  \node[anchor=west, font=\large, text=green!55!black]
    at ([xshift=2pt]outlbl.east) (passmark) {\checkmark};
  \node[anchor=west, font=\large, text=red!70!black]
    at ([xshift=3pt]passmark.east) (failmark) {$\times$};

  \begin{scope}[on background layer]
    \node[evalbox, fit=(evalcard1)(evalcard2)(outlbl)(passmark)(failmark),
          inner xsep=10pt, inner ysep=6pt] (evalbox) {};
  \end{scope}
  \node[docker title, fill=blue!45!black, text=white,
        rounded corners=2pt, inner xsep=6pt, inner ysep=2pt,
        anchor=south, yshift=3pt]
    at (evalbox.north) {Evaluation};

  \end{tikzpicture}%
  }
  \caption{\textbf{\projectname is a unified evaluation framework for
  agentic healthcare AI.} Each benchmark task is packaged as a Docker
  container environment with three stages: (1) the container sets up the environment by staging patient data and necessary artefacts under \texttt{/data/} (here, the \taskXray
  task consists of one patient's longitudinal chest-X-ray history where
  the target study's report carries a corrupted draft FINDINGS section the
  agent must correct); (2) the agent is handed a natural-language
  instruction; (3) the agent acts in a terminal, issuing tool calls (it
  \rev{uses the preinstalled tools and local commands to view, crop, and zoom into images or consult} the patient's prior
  studies) until it writes the
  corrected report to \texttt{/workspace/submission.json}.
  \rev{A separate verifier evaluates the output against hidden references} and a binary success criterion. The same containerized framework
  instantiates all \numtaskcategories task categories (right), each in a unique
  environment, with the input modality and
  number of tasks shown for every category.
  }
  \label{fig:overview}
\end{figure}

The field of AI research is undergoing a transition from large language models (LLMs) as isolated task solvers to agentic systems that can execute actions within complex environments. Traditional LLM benchmarks were largely designed around static inputs: a model receives a fixed prompt, often containing a short text context, and produces a single answer. For example, tasks such as MedQA \citep{jin2021disease} evaluate whether an LLM can answer a question given a concise clinical vignette. Nowadays, LLM agents equipped with tools, harnesses, and execution environments can go beyond fixed-context prompting: they can search, inspect, query, compute, and iteratively act on data distributed across files, databases, images, and software systems. As a result, developing agentic evaluation environments has become essential for accurately measuring the capabilities of these frontier systems.

This shift is \auditrev{especially} important for healthcare, where many high-value applications require end-to-end reasoning over complex, multimodal, and large-scale data. Facilitating real clinical workflows often involves processing information that cannot be fully represented in a short prompt, such as 3D CT volumes, gigapixel pathology slides, longitudinal EHR databases, trial protocols, and heterogeneous clinical documents. These settings have historically been beyond the reach of conventional prompting approaches, either because the data exceeds the model context window or because solving the task requires multi-step interaction with tools and the environments. Agentic systems create the possibility of addressing these workflows more realistically, by allowing models to explore the environment, operate over raw data, and design and execute task-specific strategies.

It is therefore crucial to move beyond static evaluation and develop realistic, holistic benchmark suites that assess the emerging capabilities of agents on end-to-end healthcare tasks. However, the current healthcare benchmark landscape remains fragmented and insufficient for evaluating the full capabilities of AI agents. Traditional benchmarks primarily measure isolated reasoning or question-answering ability, while recent agent-oriented efforts \citep{bedi2026healthadminbench, medagentbench, liu2026automedbench, healthbench} remain limited in scope, often focusing on specific tasks, modalities, or predefined workflows. This lack of broad, agent-native evaluation makes it difficult to draw reliable conclusions about the capabilities of frontier systems in healthcare, as real-world clinical workflows rarely rely on a single task, modality, or data source (eg. clinicians routinely integrate information from diverse inputs—such as pathology slides, CT scans, radiology reports etc. to arrive at diagnoses and treatment decisions). A comprehensive, realistic, and challenging benchmark suite is therefore needed to provide a common ground for evaluating and comparing agentic AI systems in healthcare, enabling researchers to identify strengths and areas for improvement and to track progress in this rapidly evolving field.

In this study, we introduce \projectname, a unified suite of agentic healthcare environments for evaluating AI agents. \projectname consists of \numtasks tasks across \numtaskcategories categories each with its unique environment, including \rev{\taskXray, \taskCt, \taskMri, \taskTumor, \taskSlide, \taskDq, \taskCqa, \taskEhrshot, and \taskSurvival. These tasks span \auditrev{diagnosis, specimen assessment, research, and data management workflows}.} They also cover \rev{six modalities: 2D radiographs, 3D CT volumes, 3D MRI volumes, whole-slide pathology images, free text, and structured clinical data.} As detailed in Section~\ref{sec:task_creation}, we provide a principled workflow to select and create each task in \projectname to ensure a unified benchmark that is realistic, challenging, verifiable and diverse.
For each task, we provide a terminal-based environment following \citet{merrill2026terminal} as shown in Figure~\ref{fig:overview}. Within each environment, agents are given real clinical artefacts, including patient data and supporting resources, along with minimal task instructions. This setup allows agents to freely explore the environment, formulate strategies, and propose solutions. The \taskXray task shown in Figure~\ref{fig:overview} hands the agent an environment containing a patient's longitudinal chest X-ray history and a corrupted report, and the agent can view, zoom into, and crop the radiographs and cross-reference the prior studies before writing a corrected report. In this way, \projectname evaluates agents' end-to-end autonomous capabilities, including planning, tool use, environment exploration, and task execution. For another example, when presented with a CT volume, an agent may choose to inspect slices sequentially, \rev{use preinstalled image-processing libraries}, pre-filter relevant regions, or crop candidate lesions before making a final prediction. Finally, we provide a unified evaluation module to assess agent output by defining success/failure criteria for each task \rev{against reference labels, executable rules, or computational baselines, enabling comparison of binary task outcomes across the suite.}

We use \projectname to conduct an empirical study of frontier LLM agents. We find that the tasks are challenging even for state-of-the-art systems, with the strongest agent, \rev{Claude Code Opus-5.5}, achieving $\bestResolution$ task success rate. Beyond this aggregate result, the suite enables nuanced analyses of agent strengths and weaknesses across task categories, clinical workflows, and data modalities, highlighting areas where current agents show promise as well as where substantial progress is still needed.

\subsection*{Contributions}

In summary, our contributions are as follows:
\begin{enumerate}[leftmargin=*]
\item \textbf{A unified suite of agentic healthcare environments.} We introduce \projectname, a benchmark suite for evaluating AI agents on realistic, patient-grounded tasks spanning diverse clinical workflows and data modalities, including medical imaging, free text, and structured EHR data. \auditrev{We also set up} a principled workflow for sourcing and selecting tasks for the benchmark that can be extended for future editions.

\item \textbf{A challenging benchmark far from saturation.} \projectname is designed to measure progress in frontier healthcare agents over time. Current frontier agents achieve overall low task success rate, with the strongest \rev{agent, Claude Code Opus-5.5}, reaching $\bestResolution$, leaving substantial room for future improvement.

\item \textbf{Nuanced analyses of frontier-agent capabilities.} \rev{Our benchmark reveals category-specific strengths and weaknesses: \auditrev{pathology-area selection} and EHR prediction are comparatively tractable, while exact CT/MRI interpretation and exhaustive data-quality auditing remain difficult. The effects of task combination vary across models, with more recent models such as Opus-5.5 showing no performance degradation when spawning subagents, even as task compositionality increases. Opus-5.5 achieves the highest overall success rate, while Sonnet-5.5 offers competitive performance at substantially lower cost.}

\end{enumerate}

\section{Related Work}

Healthcare has a growing body of agent-oriented benchmarks, but coverage remains fragmented across patient-data modalities and analysis workflows, motivating a unified, executable, and multi-modal evaluation spanning diverse patient-data analysis workflows. We group recent efforts into three complementary lines of work
and discuss how \projectname relates to each.

A first line of work studies how agentic strategies can improve medical
question answering. Two representative examples are
MedAgentBoard~\cite{medagentboard}, which offers a careful
comparison of LLM multi-agent collaboration against single-LLM prompting and
strong conventional methods across medical task families, and ClinicalAgent
Bench~\cite{reflectool}, which considers a broad task suite spanning
five clinical capability dimensions together with a rich clinical
toolbox. These benchmarks are primarily organized around
question-and-answer formats, in which each instance pairs a query with
pre-supplied context and the agent's tools or collaborators act as part of the
solver. \projectname is different in focus: rather than answering a
self-contained question, the agent must explore and act within a live
environment whose state it changes over many steps.

A second line of work \rev{evaluates clinical conversation or simulated interactions.} HealthBench~\cite{healthbench} and HealthBench
Professional~\cite{healthbenchpro} provide carefully constructed,
physician-authored rubrics for \rev{responses to supplied patient or clinician conversation contexts which differs from an agent interacting with a patient simulator.}
MAI-DxO~\cite{maidxo} models diagnosis as a cost-aware sequential process over
challenging NEJM cases; and AgentClinic~\cite{agentclinic} turns medical QA and
EHR cases into a simulated clinic with interacting doctor, patient, and
measurement agents. In a similar spirit, simulated-hospital environments such as
AI Hospital~\cite{aihospital}, MedAgentSim~\cite{medagentsim}, and
CP-Env~\cite{cpenv} let models role-play clinical staff across multi-step
encounters and branching care pathways. This body of work captures clinical
reasoning and communication \auditrev{especially} well. Its focus is conversational and
simulated-patient interaction, whereas \projectname asks agents to operate
directly on raw, heterogeneous patient data.

\begin{figure}[t]
  \centering
  \includegraphics[width=0.95\linewidth]{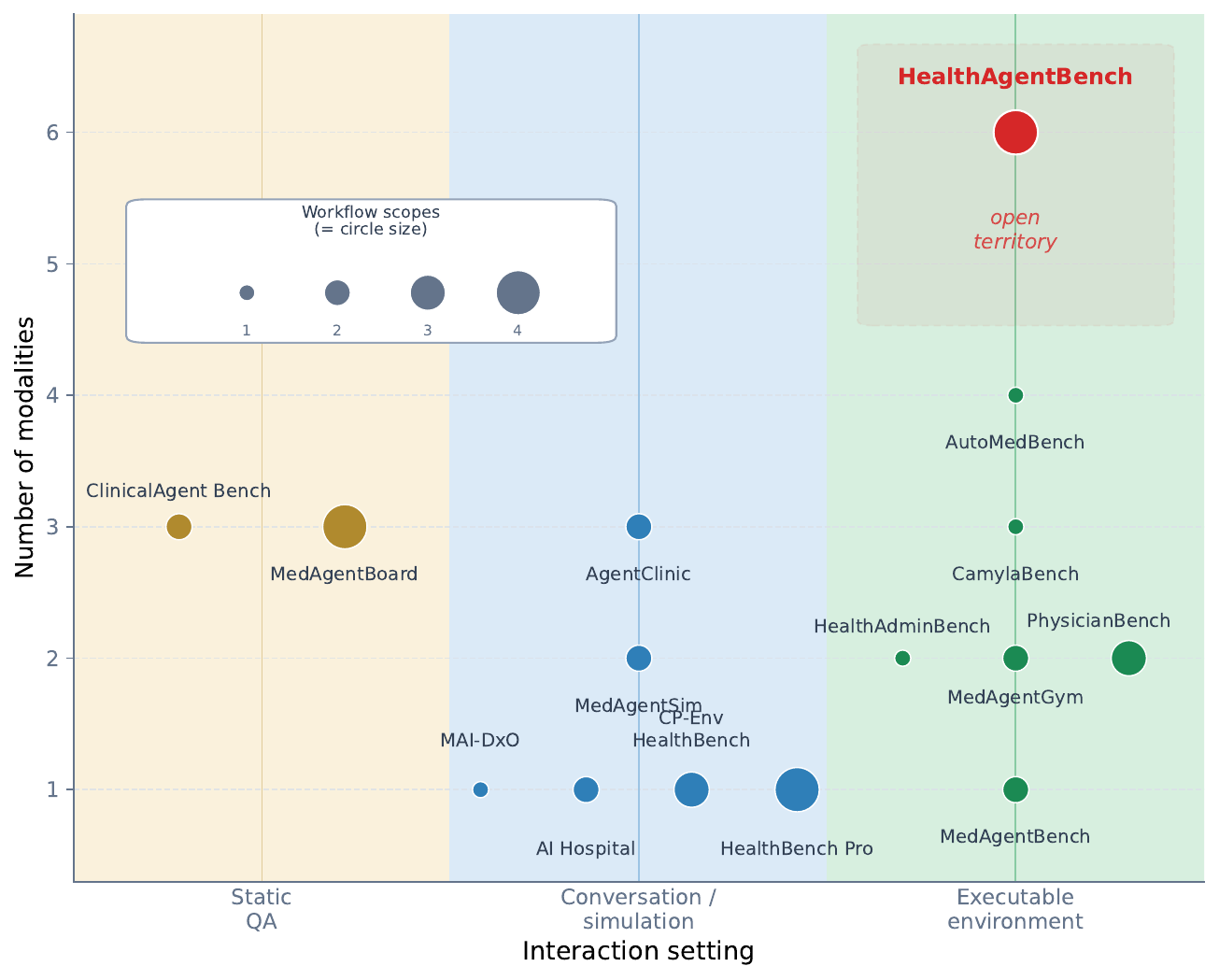}
  \caption{\textbf{Positioning of healthcare agent benchmarks.} An encoding of
  the per-benchmark comparison in Table~\ref{tab:related-work}
  (Appendix~\ref{appendix:related-work-table}). The horizontal axis is \emph{\rev{interaction setting}}, with the three interaction categories shown as
  distinctly coloured columns (static question answering,
  \rev{conversation context or simulation}, and acting in an executable
  environment); benchmarks are pinned to their column's centre, so horizontal
  position within a column carries no meaning. The vertical axis is the
  \emph{number of modalities} a benchmark covers. The \auditrev{\emph{number of clinical
  workflow scopes} is encoded as the size of each marker--a larger circle
  means more scopes (see the key, top left). \projectname (red) combines executable interaction with six modality groups and four workflow scopes: diagnosis, specimen assessment, research, and data management.}}
  \label{fig:related-work-alt}
\end{figure}

A third line of work, closest to ours, places agents in executable environments
over real data, with each benchmark concentrating on a particular clinical
setting. MedAgentBench~\cite{medagentbench} and the more recent
PhysicianBench~\cite{physicianbench} situate agents within a FHIR-based EHR to
retrieve records and place orders, with PhysicianBench extending the paradigm to
long-horizon, execution-verified workflows; both center on structured EHR data.
MedAgentGym~\cite{xu2026medagentgym} provides a scalable agentic environment for
code-centric reasoning over biomedical data-science tasks, emphasizing
programmatic analysis of structured records.
AutoMedBench~\cite{liu2026automedbench} and
CamylaBench~\cite{camyla} share our emphasis on end-to-end pipelines over raw data, \rev{with a primary focus on medical imaging. Our distinction is the broader combination of task categories and modalities.}
HealthAdminBench~\cite{bedi2026healthadminbench}, meanwhile, studies computer-use agents on
healthcare administrative workflows. 
\projectname complements these efforts by unifying their strengths into a single, broad, executable benchmark in which agents autonomously interact with real, heterogeneous patient data. By adopting terminal-based environments with minimal instructions, \projectname builds on the design philosophy of general-domain agent benchmarks such as Terminal-Bench~\cite{merrill2026terminal}, OSWorld~\cite{osworld}, GAIA~\cite{gaia}, and WebArena~\cite{webarena}, while adapting it to the multimodal data, complex workflows, and specialized tools characteristic of healthcare. Figure~\ref{fig:related-work-alt} positions \projectname relative to existing healthcare agent benchmarks along two dimensions: \rev{interaction setting} and breadth of coverage. \projectname \rev{combines executable agent interaction with coverage of six modality groups and \auditrev{four workflow groups defined by immediate task purpose.}}

\section{Task Creation\label{sec:task_creation}}

We propose a principled workflow  for selecting and designing tasks in \projectname to create \numtasks tasks across \numtaskcategories task categories including \rev{\taskXray, \taskCt, \taskMri, \taskTumor, \taskSlide, \taskDq, \taskCqa, \taskEhrshot, and \taskSurvival.}
Documentation for tasks in each task category is presented in
Appendix~\ref{appendix:benchmarks_task_categories}. Figure~\ref{fig:task-workflow} summarizes the
selection criteria, construction sources, design principles, and
post-creation checks that every task passes through. The workflow proceeds through four sequential stages: a candidate is screened against
the \emph{selection criteria}, then constructed from a vetted
\emph{source}, built under a fixed set of \emph{design principles},
and finally subjected to \emph{post-creation checks} before it joins
the suite.

\begin{figure}[t]
  \centering
  \begin{tikzpicture}[
    font=\sffamily\footnotesize,
    bullet/.style={inner sep=0pt},
    leftbox/.style={
      draw=orange!75!black, line width=0.8pt,
      fill=orange!82!black, text=white,
      rounded corners=7pt,
      inner xsep=8pt, inner ysep=6pt,
      text width=6.3cm,
      align=left,
    },
    rightbox/.style={
      draw=cyan!55!black, line width=0.8pt,
      fill=cyan!60!black, text=white,
      rounded corners=7pt,
      inner xsep=8pt, inner ysep=6pt,
      text width=6.3cm,
      align=left,
    },
    finalbox/.style={
      draw=green!50!black, line width=0.8pt,
      fill=green!55!black, text=white,
      rounded corners=4pt,
      inner xsep=8pt, inner ysep=5pt,
      text width=6.3cm,
      align=center,
      font=\sffamily\bfseries\small,
    },
    arr/.style={-{Latex[length=2.5mm,width=2mm]},
                line width=1pt, black!55},
  ]

  \node[leftbox] (box1) {%
    {\sffamily\bfseries\small1.~Task selection criteria}\\[2pt]
    \begin{itemize}[leftmargin=1.2em,itemsep=1pt,topsep=1pt,label={}]
      \item \faRobot~~Multi-step agentic execution
      \item \faUserInjured~~Realistic clinical workflows
      \item \faLayerGroup~~Diverse environment coverage
      \item \faCheckCircle~~Verifiable; low chance success
    \end{itemize}
  };

  \node[leftbox, below=0.35cm of box1] (box2) {%
    {\sffamily\bfseries\small2.~Sourcing tasks}\\[2pt]
    \begin{itemize}[leftmargin=1.2em,itemsep=1pt,topsep=1pt,label={}]
      \item \faRecycle~~Re-design from existing benchmarks
      \item \faUserMd~~Create new tasks via injection rules
    \item \faUserCheck~~(+ manual review)

    \end{itemize}
  };

  \node[leftbox, below=0.35cm of box2] (box3) {%
    {\sffamily\bfseries\small3.~Task creation principles}\\[2pt]
    \begin{itemize}[leftmargin=1.2em,itemsep=1pt,topsep=1pt,label={}]
      \item \faTerminal~~Terminal envs for any raw data format
      \item \faFilter~~Efficient, representative sampling
      \item \faDownload~~\rev{Task-specific public/gated data staging}
      \item \faLock~~Anti-cheat by construction
      \item \faFeather~~\rev{Minimal instructions; autonomous strategy selection}
      \item \faCheckDouble~~Determine success criteria
    \end{itemize}
  };

  \node[rightbox, right=1.4cm of box2] (box4) {%
    {\sffamily\bfseries\small4.~Post-creation checks}\\[2pt]
    \begin{itemize}[leftmargin=1.2em,itemsep=1pt,topsep=1pt,label={}]
      \item \faSearch~~Manual + LLM review of trajectories
      \item \faRandom~~\rev{Check ambiguity and unintended shortcuts}
      \item \faTrash~~\auditrev{Discard trivially solved categories}
    \end{itemize}
  };

  \node[finalbox, below=0.7cm of box4] (box5)
    {\faCheckCircle~~Task added to \projectname};

  \draw[arr] (box1.south) -- (box2.north);
  \draw[arr] (box2.south) -- (box3.north);
  \draw[arr] (box3.east) to[out=0,in=180] (box4.west);
  \draw[arr] (box4.south) -- (box5.north);

  \end{tikzpicture}
  \caption{\textbf{The \projectname task-creation workflow.} Each candidate task progresses through four stages: (1) satisfying the selection criteria, (2) construction from either an existing benchmark or curated raw patient data, (3) application of standardized task design principles, and (4) \rev{post-creation checks for unintended shortcuts, \auditrev{ambiguity, and trivially solved task categories before inclusion}.}}
  \label{fig:task-workflow}
\end{figure}

\subsection{Task selection criteria.} To be included in \projectname, a candidate task must satisfy the following criteria 
\paragraph{Agentic Workflow} We deliberately choose tasks that require an \emph{agentic} workflow that can involve planning, tool use, multi-step reasoning, or environment interaction that naive single-shot LLM prompting cannot complete. For example, the LLM cannot pass the ct volume, or pathology slide as a whole into the prompt.

\paragraph{Realistic Clinical Workflow} \auditrev{We group tasks into four workflows by their primary purpose. \emph{Diagnosis} involves interpreting clinical evidence to identify abnormalities or assess disease, represented by \taskXray, \taskCt, and \taskMri. \emph{Specimen assessment} involves identifying and selecting tissue specimens or regions relevant to further analysis, represented by \taskSlide and \taskTumor. \emph{Research} broadly encompasses evidence synthesis, scientific data analysis, experimentation, and model development and evaluation, consistent with the related-work inventory in Table~\ref{tab:related-work}. Our suite represents this scope through \taskEhrshot and \taskSurvival, which require developing predictive models to support patient-event monitoring and prognosis. \emph{Data management} involves retrieving, organizing, transforming, or auditing patient data as the task's primary objective, represented by \taskCqa and \taskDq. These workflows can overlap in clinical practice. All tasks use real patient records or images, with task-specific synthetic modifications where appropriate.}

\paragraph{Wide coverage of modalities and environments} 
Our task suite is designed to span diverse healthcare settings and environments. Figure~\ref{fig:coverage-matrix} summarizes the coverage of \projectname across multiple dimensions, including input modality, clinical workflow, output format, patient and temporal scope, and data access regime. \projectname spans \rev{six input modalities---2-D radiographs, 3-D CT volumes, 3-D MRI volumes, pathology whole-slide images, free-text clinical documents, and structured EHR data.} Its tasks also cover diverse agent output formats, including prose reports, \rev{numerical answers, classifications, selected image regions/sets, predictions, and flagged data-quality rows.} The suite also varies patient scope, from single-record reasoning to cohort-scale analysis, and includes longitudinal tasks that require reasoning over time-ordered events.

\paragraph{Verifiable with low chance success rate}\rev{
We require clear, verifiable pass criteria and aim to minimize success by random guessing. A single balanced binary classification question, for example, can be answered correctly by chance with probability $50\%$. We therefore bundle multiple questions into a single task and require all answers to be correct to pass. Requiring multiple correct decisions makes successful completion more informative about the agent's capabilities.}

\subsection{Sourcing tasks.} Candidate tasks are constructed from two sources. First, we leverage existing benchmarks with expert labels or reports and convert them into terminal-based agent environments. We redesign the agent-visible interface so that agents must solve the tasks end-to-end through autonomous exploration, without the human scaffolding or task-specific engineering used in the original benchmark evaluations. \rev{For example, \taskCqa adapts the compositional information-seeking component of CodeClinic~\citep{ossowski2026codeclinic}.} \taskEhrshot is adapted from EHRSHOT~\citep{ehrshot}. Whereas EHRSHOT was originally designed for human researchers to develop predictive models, our version requires the agent to perform the entire machine learning workflow from exploring the patient database and selecting a modeling strategy to training a model and generating predictions on the test set, thereby transforming it into a substantially more challenging agentic evaluation. \auditrev{Second, we create new tasks} by augmenting curated patient datasets or wrapping existing clinical pipelines into executable environments. For example, \taskDq is built by injecting realistic data quality issues into the MIMIC-IV dataset using rule-based perturbations, such as impossible physiological values inconsistent clinical records and demographic conflicts. The agent must inspect the underlying patient data, identify the injected errors, and report the corresponding data quality issues.

\subsection{Task design principles.} Every task is built around
\rev{six} design principles as described below:
\paragraph{\emph{(i) Versatile terminal environment.}}
We leverage the harbor framework \citep{Harbor_Framework_Team_Harbor_A_framework_2026} to package each task as a terminal environment that accommodates
the heterogeneous raw patient-data formats the suite spans. Terminal environments provide a flexible interface through which agents can inspect heterogeneous data formats, invoke domain-specific software, query databases, and compose complex workflows over large-scale resources. They also closely match the execution environment of today's frontier coding agents, which are primarily designed and optimized to operate through terminal interfaces. This makes terminal environments a natural substrate for evaluating autonomous agent capabilities on realistic clinical tasks.\footnote{Our choice of a terminal interface is motivated by evaluation rather than deployment. In practical clinical systems, agents could operate in the backend through a terminal or execution environment while presenting their outputs through clinician-facing graphical interfaces.}. 
\paragraph{\emph{(ii) Efficient, representative
sampling.}} Rather than shipping hundreds of similar tasks, we
sample a small set (typically 5--15 samples per task category) that are representative (eg. covering different disease types, difficulty levels) so a sweep of all tasks in this benchmark produces an informative signal in reasonable time. \paragraph{\emph{\rev{(iii) Task-specific data staging.}}} \rev{The suite combines credentialed bootstrap downloads, public downloads, and packaged fixed public-demo artifacts. Data delivery follows each source's access requirements and redistribution rule.} \paragraph{\emph{(iv) Anti-cheat by
construction.}} Gold labels and verifier code are mounted only into
the verifier step, never into the agent container. \rev{Where applicable, source identifiers are made opaque} (e.g.\ \texttt{case\_NN},
\texttt{study\_NN\_<timestamp>}) \rev{and source names are scrubbed from imaging filenames} \rev{to reduce direct source lookup. Agent web tools are disabled and \auditrev{runtime internet access is restricted to model APIs}.} \paragraph{\emph{(v) Minimalist instructions.}} The
agent-facing prompt \rev{defines the task contract, resources, and supplied training examples where present, but leaves the strategy} (which files to read, in what
order, which tools to call) entirely to the agent rather than predefining workflows. This way, we can test the innate planning and strategy formulation capabilities of the agents alongside their execution abilities. 
\paragraph{\emph{(vi) Determine task-specific success criteria}} Each task is evaluated using a binary success/failure criterion, with task-specific performance metrics reported alongside. The success criterion is determined for each task category separately; Table~\ref{tab:success-criteria} in Appendix~\ref{appendix:benchmarks} summarizes these criteria, and the \auditrev{per-task-category \emph{Scoring} \auditrev{paragraphs in Appendix~\ref{appendix:benchmarks_task_categories}}} give the full details. These criteria fall into a few broad types: \emph{exact or complete correctness} (\auditrev{e.g., all CT abnormality diseases are detected and labelled correctly for a CT volume}), \emph{exhaustive recovery} of the targets (eg. full recall on \rev{required data-quality clusters}), \emph{error-free output} (zero clinically-significant errors for report correction), and \emph{reaching a performance threshold}---either an absolute metric cutoff (tumor tile-F1 $\geq 0.90$) or matching a human-engineered baseline (AUROC at or above the human engineered count+GBM model). This design enables a uniform evaluation framework across the diverse tasks in the suite. We use overall task success rate (i.e. the fraction of benchmark tasks successfully resolved by an agent) as the primary metric for measuring overall agent progress and for comparing between tasks. Task-specific metrics, such as F1 and recall, are reported alongside task success rate to quantify partial progress and provide additional insight into agent performance.

\subsection{Post-creation checks.} Before a task is added to the
suite, we run a baseline sweep across multiple frontier agents and
inspect the trajectories. We \emph{manually and with LLM
assistance} review the agent transcripts to catch two failure
modes: (i)~the agent finding an unintended shortcut (e.g.
a leaked evaluation script into agent's container) which sends the task back
for redesign; and (ii)~genuine ambiguity in the task where multiple
\emph{reasonable} agent solutions or alternative valid gold
labels are possible, which also sends the task back. \auditrev{A final filter is applied: we \emph{discard trivially solved task categories}, defined as categories in which every task is passed by all evaluated agents. Individual tasks passed by all agents can remain within a retained category.}

\section{Baselines and Results}
\label{sec:baselines}

\subsection{Overall Performance}

\rev{We evaluate six agents: GPT-5.6-sol, GPT-6-astra, and GPT-6-sol through Codex~\citep{openai2025codexcli}, and Opus-5, Opus-5.5, and Sonnet-5.5 through Claude Code~\citep{anthropic2025claudecode}. Each uses xhigh reasoning effort, disabled web tools, and agent network access restricted to model APIs. We retain one scored attempt for each of 75 tasks, totaling 450 evaluations.}

\begin{figure}[t]
  \centering
  \includegraphics[width=0.95\linewidth]{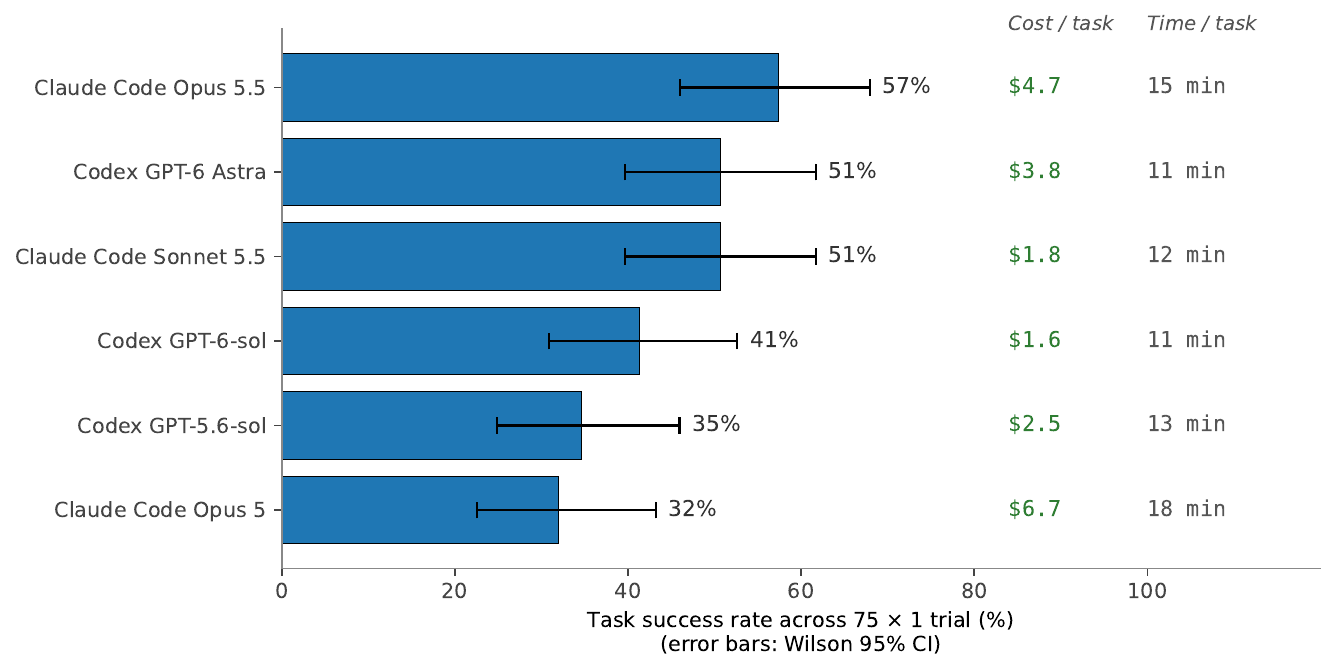}
  \caption{\rev{\textbf{Pooled task success rate of six frontier agents on \projectname.} Each bar pools one attempt on each of 75 tasks. Error bars are descriptive Wilson 95\% intervals over task outcomes. The right-hand columns show mean API-equivalent cost and complete trial wall-clock time per task.}}
  \label{fig:hero}
\end{figure}

\rev{Figure~\ref{fig:hero} summarizes overall success. Opus-5.5 leads at 43/75 (57.3\%), followed by GPT-6-astra and Sonnet-5.5 at 38/75 each (50.7\%), GPT-6-sol at 31/75 (41.3\%), GPT-5.6-sol at 26/75 (34.7\%), and Opus-5 at 24/75 (32.0\%). Overall, the benchmark reveals a clear and steady improvement across successive model generations while remaining far from saturation.}

\rev{Beyond success, we report API-equivalent agent cost and complete trial wall-clock time. The observed cost--performance frontier comprises GPT-6-sol (\$119.65 per sweep), Sonnet-5.5 (\$136.15), and Opus-5.5 (\$353.89; Figure~\ref{fig:cost-performance}). Sonnet-5.5 is cost-effective as it shows competitive performance with \auditrev{about 38\% of Opus-5.5's cost}.}

\subsection{Breaking Down Performance by Task Categories}

\begin{figure}[t]
  \centering
  \includegraphics[width=\linewidth]{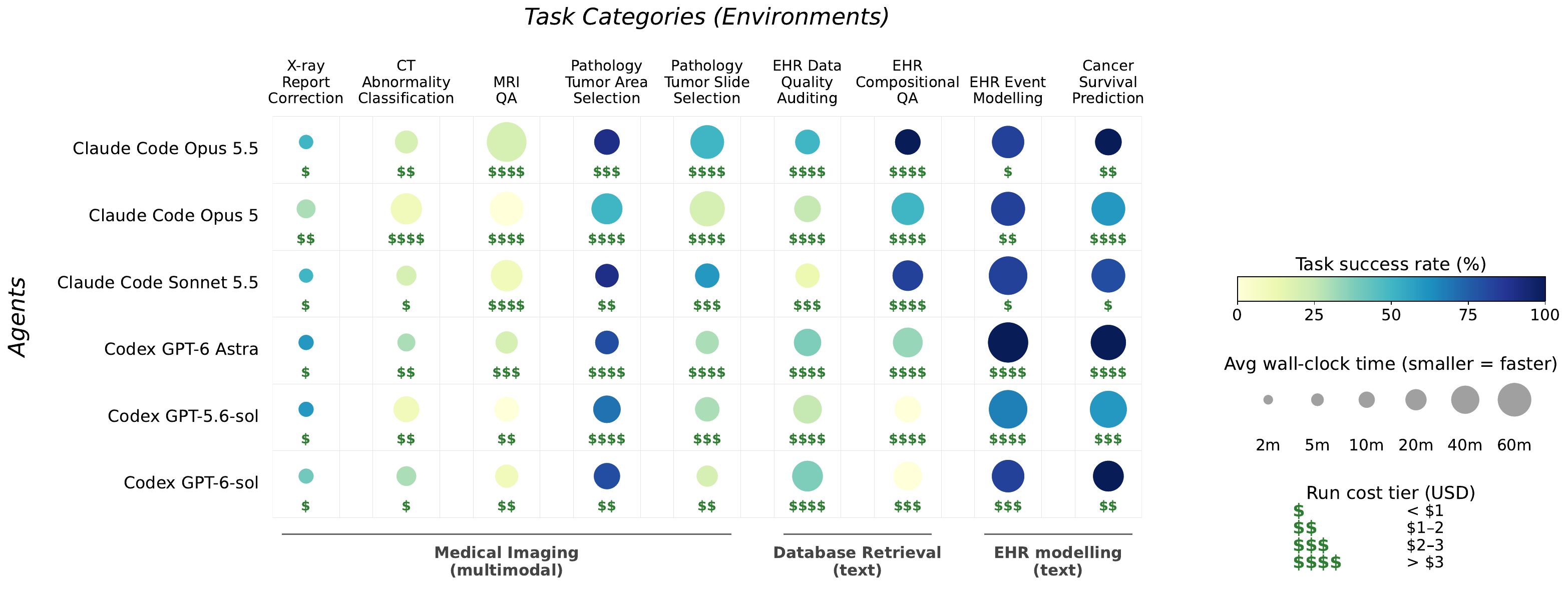}
  \caption{\rev{\textbf{Performance and efficiency of six agents across nine task categories.} Color denotes category task success rate; circle size is mean complete trial wall-clock time; \$ markers encode mean per-trial agent cost. Each task contributes one scored attempt. Exact values appear in Tables~\ref{tab:per-cat-success}--\ref{tab:per-cat-cost}; partial metrics appear in Figure~\ref{fig:task-specific}.}}
  \label{fig:baseline-heatcircle}
\end{figure}

\rev{Figure~\ref{fig:baseline-heatcircle} decomposes performance by task category, encoding success rate, complete trial wall-clock time, and cost. No agent is uniformly best across categories. Task-specific raw scores in Figure~\ref{fig:task-specific} quantify partial progress that binary task rewards can obscure.}

\paragraph{\rev{The Win: Building Predictive Models}}

\rev{\taskEhrshot and \taskSurvival assess agents' ability to build predictive models from raw longitudinal EHRs and clinical/genomic training tables, respectively. Agents must process the data, select and fit models, and generate predictions for held-out patients. Success requires meeting or exceeding the Count+GBM reference on AUROC for event modelling, or outperforming both Cox and RSF references on all three survival metrics. Across these 11 tasks, every agent passes at least 7 (63.6\%), and GPT-6-astra passes all 11. These results show that agents can build prediction pipelines competitive with the specified human-engineered references under the benchmark's resource limits. Reviewed successful runs engineer temporal or clinical features and compare candidate models on validation data; survival strategies also include ensembling and time calibration. Failed EHR runs fall below the AUROC reference, while failed survival runs miss the IBS gate, sometimes despite favorable validation scores. Appendix~\ref{appendix:trajectories} illustrates these strategies and performance shortfalls with successful and failed attempts. Two GPT-5.6-sol runs also contain small metric-implementation mistakes, but their frozen predictions fail under both local and official metrics, so these mistakes alone do not explain the failures (Appendix~\ref{appendix:bench:survival}).}

\rev{These success rates measure performance relative to reference pipelines; they do not establish that clinical event or survival prediction is solved. The continuous metrics in Figures~\ref{fig:ehr-auroc} and~\ref{fig:survival-tte} reveal differences in discrimination, prediction error, and calibration that binary outcomes obscure. The attainable performance ceiling given the available patient records is unknown: incomplete information and outcome uncertainty may limit prediction accuracy. Exact agreement with every patient's observed event timeline is therefore not an appropriate standard for declaring these prediction problems solved.}

\paragraph{\rev{The Challenge: Exhaustive Retrieval from a Complex Search Space}}

\rev{Both \taskDq and \taskCqa require agents to retrieve and reason over EHR data. \taskCqa provides specific clinical questions that guide queries and calculations, whereas \taskDq requires agents to discover errors across eight tables, with more than 800,000 rows in the combined package. Agents must plan the audit, prioritize promising checks, and verify findings across a large search space. Opus-5.5 leads both categories, passing 4/8 data-quality tasks (50\%) and all 6/6 compositional-QA tasks (100\%). Its reviewed data-quality runs inspect value ranges, units, duplicate records, and demographic and cross-table consistency; its CQA runs use supplied training examples to resolve calculation conventions and implement the corresponding queries. The failed examples in Appendix~\ref{appendix:trajectories} illustrate incomplete audit coverage and unimplemented QA calculations, respectively. Other reviewed CQA errors involve join or time-window conventions (Appendix~\ref{appendix:bench:cqa}).}

\rev{Combining subtasks has model-dependent effects (Figure~\ref{fig:cqa-single-combined}). In compositional QA, five individual category tasks partition the same 51 questions presented in the combined task. Opus-5.5 answers all 51 correctly in both settings. Opus-5 drops from 48 to 27 correct answers in the combined setting, and GPT-5.6-sol drops from 41 to 37. The combined-task trajectories illustrate contrasting work strategies: Opus-5.5 and Sonnet-5.5 delegate calculations to four and six internal subagents, respectively, and both achieve 51/51, whereas Opus-5 leaves 24 calculations unimplemented and fills those answers with the most frequent training-set responses without launching subagents. This result indicates that the newer agents are likely to be more capable of multi-task execution with more capable subagent parallelization.}

\paragraph{\rev{The Challenge: Heterogeneous Medical Imaging Performance}}
\rev{Medical imaging performance varies sharply by task. Opus-5.5 and Sonnet-5.5 each pass 9/10 pathology-area tasks, while the highest CT success is 3/10 (GPT-6-astra and GPT-6-sol), and the highest MRI success is 2/10 (GPT-6-astra and Opus-5.5). Reviewed pathology successes combine whole-slide overviews with higher-resolution crops to localize tumor tissue, demonstrating strategies for navigating large images and converting visual findings into tile selections.} \rev{Reviewed CT and MRI runs load volumes and inspect rendered slices or crops, yet still disagree with reference labels after targeted inspection. These trajectories suggest difficulty recognizing subtle anatomical abnormalities and integrating evidence across slices and imaging planes. Image presentation and context management may also contribute, so the observed failures cannot be attributed solely to visual reasoning (Appendix~\ref{appendix:trajectories}).}

\section{Conclusion}

We introduced \projectname, a unified, agent-native benchmark suite for evaluating healthcare AI agents across diverse data modalities and \rev{\auditrev{diagnosis, specimen assessment, research, and data management workflows}}. The benchmark remains highly challenging for current frontier agents: the strongest agent, \rev{Claude Code Opus-5.5}, achieves an \rev{overall} success rate of \bestResolution across the suite, leaving substantial room for future progress. \rev{Our evaluation of six agents reveals strong results in \auditrev{pathology-area selection} and predictive modelling alongside persistent difficulty in specialized 3D imaging modalities such as CT/MRI interpretation and exhaustive auditing. Some newer agents maintain or improve accuracy on combined compositional QA tasks, with successful trajectories illustrating effective task decomposition and subagent coordination. }
While \projectname offers a practical interface for measuring and tracking progress
in agentic AI for healthcare, we acknowledge that the benchmark is not exhaustive for all aspects of healthcare agents, and we encourage future work to add additional tasks and modalities to better capture the full spectrum of agentic AI capabilities in healthcare, and we hope it becomes a useful community resource for
evaluating future healthcare agents.

\newpage
\bibliographystyle{plainnat}  %
\bibliography{refs}

\newpage
\appendix

\section{Benchmark Comparison}
\label{appendix:related-work-table}
Table~\ref{tab:related-work} gives the full per-benchmark comparison from which
the positioning of Figure~\ref{fig:related-work-alt} is derived.
\begin{table}[!ht]
  \centering
  \footnotesize
  \setlength{\tabcolsep}{5.5pt}
  \renewcommand{\arraystretch}{1.15}
  \resizebox{\textwidth}{!}{%
  \begin{tabular}{@{}lllp{5.2cm}r@{}}
    \toprule
    \textbf{Benchmark} & \textbf{Modality} & \textbf{Interaction} &
      \auditrev{\textbf{Workflow scopes}} & \auditrev{\textbf{Count}} \\
    \midrule
    \multicolumn{5}{@{}l}{\emph{Question answering}}\\
    MedAgentBoard~\cite{medagentboard}        & text, EHR, 2D  & static QA        & \auditrev{diagnosis, comms, data, research} & \auditrev{4} \\
    ClinicalAgent Bench~\cite{reflectool}     & text, EHR, 2D  & static QA        & \auditrev{diagnosis, data} & \auditrev{2} \\
    \addlinespace[2pt]
    \multicolumn{5}{@{}l}{\emph{\rev{Conversation context / simulated clinic}}}\\
    HealthBench~\cite{healthbench}            & text           & \rev{conversation context} & \auditrev{diagnosis, treatment, comms} & \auditrev{3} \\
    HealthBench Professional~\cite{healthbenchpro} & text      & \rev{conversation context} & \auditrev{diagnosis, treatment, comms, research} & \auditrev{4} \\
    MAI-DxO~\cite{maidxo}                     & text           & dialogue / simulation & \auditrev{diagnosis} & \auditrev{1} \\
    AgentClinic~\cite{agentclinic}            & text, EHR, 2D  & dialogue / simulation & \auditrev{diagnosis, comms} & \auditrev{2} \\
    AI Hospital~\cite{aihospital}             & text           & dialogue / simulation & \auditrev{diagnosis, comms} & \auditrev{2} \\
    MedAgentSim~\cite{medagentsim}            & text, 2D       & dialogue / simulation & \auditrev{diagnosis, comms} & \auditrev{2} \\
    CP-Env~\cite{cpenv}                       & text           & dialogue / simulation & \auditrev{diagnosis, treatment, comms} & \auditrev{3} \\
    \addlinespace[2pt]
    \multicolumn{5}{@{}l}{\emph{Executable environment}}\\
    MedAgentBench~\cite{medagentbench}        & EHR            & executable     & \auditrev{treatment, data} & \auditrev{2} \\
    PhysicianBench~\cite{physicianbench}      & text, EHR      & executable     & \auditrev{diagnosis, treatment, comms} & \auditrev{3} \\
    MedAgentGym~\cite{xu2026medagentgym}      & text, EHR      & executable     & \auditrev{data, research} & \auditrev{2} \\
    AutoMedBench~\cite{liu2026automedbench}   & \rev{2D, CT, MRI, WSI} & executable     & \auditrev{research} & \auditrev{1} \\
    CamylaBench~\cite{camyla}                 & \rev{2D, CT, MRI} & executable     & \auditrev{research} & \auditrev{1} \\
    HealthAdminBench~\cite{bedi2026healthadminbench} & GUI, EHR & executable   & \auditrev{admin} & \auditrev{1} \\
    \midrule
    \textbf{\projectname} & \rev{\textbf{text, EHR, 2D, CT, MRI, WSI}} & \textbf{executable} &
      \auditrev{\textbf{diagnosis, specimen assessment, research, data management}} & \auditrev{\textbf{4}} \\
    \bottomrule
  \end{tabular}%
  }
  \caption{\textbf{Where \projectname sits among healthcare agent benchmarks.}
  \projectname
  row spans \rev{six modalities and \auditrev{four workflow scopes}. CT and MRI count separately for every benchmark.} The positioning of
  Figure~\ref{fig:related-work-alt} is read directly off this table.
  \emph{Modality} counts data the agent must \emph{process as raw input}:
  \textbf{text}; \textbf{EHR} (structured/tabular records via FHIR/SQL);
  \textbf{2D} (2-D image---radiograph, dermatology/photo, ultrasound, fundus,
  blood smear); \rev{\textbf{CT} (CT volume); \textbf{MRI} (MRI volume)}; \textbf{WSI} (whole-slide
  pathology); \textbf{GUI} (computer-use web interface).
  \emph{Interaction} is the \rev{descriptive interaction setting}, the
  $x$-axis of Figure~\ref{fig:related-work-alt}: \textbf{static QA},
  \rev{\textbf{conversation context / simulation}}, or \textbf{executable} environment.
  \auditrev{We assign scopes using seven shared, overlapping labels:
  \textbf{diagnosis} (diagnostic interpretation), \textbf{specimen} (specimen assessment),
  \textbf{treatment} (treatment and care planning or execution),
  \textbf{comms} (communication and documentation), \textbf{data} (data management),
  \textbf{research} (evidence synthesis, scientific data analysis, experimentation, and model development/evaluation), and \textbf{admin} (healthcare administration).
  Counts summarize our mapping of principal evaluated scopes, not independent workflow stages or a calibrated breadth score. In PhysicianBench, EHR retrieval supports clinical care and is not counted as a separate data-management workflow. HealthBench is classified by its clinical-advice and communication workflows; isolated record-review or literature questions do not constitute separate data-management or research workflows. HealthBench Professional explicitly includes medical research assistance, distinct from autonomous experimentation or model development.
  For \projectname, the four workflow groups match Figure~\ref{fig:coverage-matrix};
  research includes event-prediction and survival-model development.}}
  \label{tab:related-work}
\end{table}

\section{The \projectname Benchmark Suite}
\label{appendix:benchmarks}

\sloppy  %

\projectname is built around the Harbor execution substrate~\citep{Harbor_Framework_Team_Harbor_A_framework_2026}: each task is
packaged as a Harbor task with \rev{a standardized container, preinstalled analysis tools, agent-visible instructions, hidden test labels, and a separate verifier. Build/bootstrap services stage public or credentialed data; agent execution permits model API access only, and verifiers are offline except for X-ray judge API calls. The current suite includes nine categories: \taskXray, \taskCt, \taskMri, \taskTumor, \taskSlide, \taskDq, \taskCqa, \taskEhrshot, and \taskSurvival. Table~\ref{tab:bench-overview} summarizes the inventory.}
\begin{table}[!ht]
\centering
\small

\begin{tabular}{p{4.3cm}p{2.7cm}p{4.0cm}r}
\toprule
Task Category & Modality & Task type & \# tasks \\
\midrule
\taskXray & Imaging + text & Report correction & 10 \\
\taskCt & 3D CT & Abnormality classification & 10 \\
\taskMri & 3D MRI & ACL/meniscus QA & 10 \\
\taskTumor & Pathology WSI & Tumor area selection & 10 \\
\taskSlide & Pathology WSI & Tumor slide selection & 10 \\
\taskDq & Tabular EHR & Data-quality auditing & 8 \\
\taskCqa~\citep{ossowski2026codeclinic} & EHR + text & Clinical calculations & 6 \\
\taskEhrshot & Longitudinal EHR & Event prediction & 6 \\
\taskSurvival & Clinical/genomic tables & Survival-time prediction & 5 \\
\bottomrule
\end{tabular}
\caption{\rev{Task categories in the current 75-task suite.}}
\label{tab:bench-overview}
\end{table}

\paragraph{Success criteria.}
Every task in \projectname reduces to a single binary success/failure outcome:
a trial earns reward~$1$ only when it meets the task's success criterion, and
$0$ otherwise. The criteria are defined per task category based on different
underlying metrics (verifier checks, error counts, accuracy, F1, recall, \rev{AUROC, and survival metrics}). The task success rates reported in Section~\ref{sec:baselines} aggregate these binary outcomes across tasks.
Table~\ref{tab:success-criteria} summarizes the success criteria for every task
category; the \auditrev{per-task-category \emph{Scoring} \auditrev{paragraphs in Appendix~\ref{appendix:benchmarks_task_categories}}} give the
full details, and the per-task fine-grained metrics are reported in
Appendix~\ref{appendix:per-task}.

\begin{table}[!ht]
\centering
\small

\begin{tabular}{@{}p{4.1cm}p{2.8cm}p{7.0cm}@{}}
\toprule
Task Category & Success metric & Success criteria (reward $=1$) \\
\midrule
\taskXray & Significant errors & At least three saved judge votes find zero significant errors; five calls intended \\
\taskCt & Label accuracy & Every requested retained label correct \\
\taskMri & Answer accuracy & All six ACL/meniscus answers correct \\
\taskTumor & Tile-level F1 & Tumor-tile F1 $\geq 0.90$ \\
\taskSlide & Exact slide set & Predicted tumor-slide set equals reference set \\
\taskDq & Cluster recall & All required injected clusters found and row precision $\geq 0.1$ \\
\taskCqa & Answer accuracy & Every answer correct within its specified tolerance; valid submission \\
\taskEhrshot & Test AUROC & AUROC $\geq$ the Count+GBM reference, with zero margin \\
\taskSurvival & C-index, cMAE, IBS & Full coverage; C-index strictly above, and cMAE and IBS strictly below, both Cox and RSF references \\
\bottomrule
\end{tabular}
\caption{\rev{Executable success criteria for all nine task categories. Partial metrics accompany the binary reward in Figure~\ref{fig:task-specific}.}}
\label{tab:success-criteria}
\end{table}

\paragraph{Coverage across evaluation axes.}
Beyond the per-task-category summary of Table~\ref{tab:bench-overview},
\rev{Figure~\ref{fig:coverage-matrix} maps the nine categories onto five evaluation axes: modality (2D radiographs, 3D CT, 3D MRI, pathology whole slides, free text, structured EHR), \auditrev{workflow (diagnosis, specimen assessment, research, data management)}, output shape (prose, classification, numerical/solution), patient/temporal scope, and data access. The summary row reports the union of observed coverage, rather than asserting coverage of every possible healthcare workflow.}

\def\countXray{10}
\def\countCt{10}
\def\countMri{10}
\def\countTumor{10}
\def\countSlide{10}
\def\countCqa{6}
\def\countDq{8}
\def\countEhrshot{6}
\def\countTte{5}
\def\countTotal{75}

\begin{figure}[tbp]
  \centering
  \begin{tikzpicture}[
    font=\sffamily\scriptsize,
    x=0.45cm, y=0.45cm,
  ]

  \definecolor{cMod}{HTML}{1F77B4}
  \definecolor{cWf} {HTML}{2CA02C}
  \definecolor{cOut}{HTML}{D62728}
  \definecolor{cScp}{HTML}{9467BD}
  \definecolor{cAcc}{HTML}{FF7F0E}
  \definecolor{cCre}{HTML}{8C564B}  %
  \definecolor{cSub}{HTML}{555555}  %

  \def\xMa{0} \def\xMb{1} \def\xMf{2} \def\xMc{3} \def\xMd{4} \def\xMe{5}
  \def\xWa{6.4} \def\xWb{7.4} \def\xWc{9.4} \def\xWd{8.4}
  \def\xOa{10.8} \def\xOc{11.8} \def\xOd{12.8}
  \def\xPa{14.2} \def\xPb{15.2} \def\xPc{16.2}
  \def\xAa{17.6} \def\xAb{18.6}
  \def\xV{20.0}

  \node[rotate=90, anchor=west, text=cMod] at (\xMa+0.5, 0.1) {Longitudinal X-ray};
  \node[rotate=90, anchor=west, text=cMod] at (\xMb+0.5, 0.1) {\rev{3D CT}};
  \node[rotate=90, anchor=west, text=cMod] at (\xMf+0.5, 0.1) {\rev{3D MRI}};
  \node[rotate=90, anchor=west, text=cMod] at (\xMc+0.5, 0.1) {WSI};
  \node[rotate=90, anchor=west, text=cMod] at (\xMd+0.5, 0.1) {Free Text};
  \node[rotate=90, anchor=west, text=cMod] at (\xMe+0.5, 0.1) {Structured EHR};

  \node[rotate=90, anchor=west, text=cWf, align=left] at (\xWa+0.5, 0.1) {Diagnosis};
  \node[rotate=90, anchor=west, text=cWf, align=left] at (\xWb+0.5, 0.1) {Specimen assessment};
  \node[rotate=90, anchor=west, text=cWf, align=left] at (\xWc+0.5, 0.1) {Data management};
  \node[rotate=90, anchor=west, text=cWf, align=left] at (\xWd+0.5, 0.1) {Research};

  \node[rotate=90, anchor=west, text=cOut] at (\xOa+0.5, 0.1) {Prose};
  \node[rotate=90, anchor=west, text=cOut] at (\xOc+0.5, 0.1) {Classification};
  \node[rotate=90, anchor=west, text=cOut] at (\xOd+0.5, 0.1) {\rev{Numerical / solution}};

  \node[rotate=90, anchor=west, text=cScp] at (\xPa+0.5, 0.1) {Single-patient};
  \node[rotate=90, anchor=west, text=cScp] at (\xPb+0.5, 0.1) {Cohort};
  \node[rotate=90, anchor=west, text=cScp] at (\xPc+0.5, 0.1) {Longitudinal};

  \node[rotate=90, anchor=west, text=cAcc] at (\xAa+0.5, 0.1) {Public};
  \node[rotate=90, anchor=west, text=cAcc] at (\xAb+0.5, 0.1) {Gated};

  \node[rotate=90, anchor=west, text=cSub] at (\xV+0.5, 0.1) {\# tasks};

  \newcommand{\groupbar}[4]{%
    \draw[#4, line width=0.7pt]
      (#1, 6.9) -- (#1, 6.7) -- (#2, 6.7) -- (#2, 6.9);
    \node[text=#4, anchor=south, font=\sffamily\tiny\bfseries]
      at ({(#1+#2)/2}, 6.85) {#3};
  }
  \groupbar{\xMa}{\xMe+1}{Modality}{cMod}
  \groupbar{\xWa}{\xWc+1}{Workflow}{cWf}
  \groupbar{\xOa}{\xOd+1}{Output}{cOut}
  \groupbar{\xPa}{\xPc+1}{Scope}{cScp}
  \groupbar{\xAa}{\xAb+1}{Access}{cAcc}
  \groupbar{\xV}{\xV+1}{Count}{cSub}

  \newcommand{\rowlabel}[2]{%
    \node[anchor=east, font=\sffamily\scriptsize]
      at (-0.3, -#1 + 0.5) {#2};
  }
  \rowlabel{1}{\rev{\taskMri}}
  \rowlabel{2}{\taskXray}
  \rowlabel{3}{\rev{\taskSlide}}
  \rowlabel{4}{\taskCt}
  \rowlabel{5}{\taskTumor}
  \rowlabel{6}{\taskEhrshot}
  \rowlabel{7}{\taskDq}
  \rowlabel{8}{\rev{\taskCqa}}
  \rowlabel{9}{\rev{\taskSurvival}}
  \node[anchor=east, font=\sffamily\scriptsize\bfseries]
    at (-0.3, -10 + 0.5) {\rev{ALL tasks}};

  \foreach \xcol in {\xMa,\xMb,\xMf,\xMc,\xMd,\xMe,%
                     \xWa,\xWb,\xWc,\xWd,%
                     \xOa,\xOc,\xOd,%
                     \xPa,\xPb,\xPc,%
                     \xAa,\xAb,%
                     \xV} {
    \foreach \r in {1,...,10} {
      \draw[gray!30, line width=0.3pt, fill=gray!8]
        (\xcol, -\r) rectangle ++(1, 1);
    }
  }

  \newcommand{\fcell}[3]{%
    \fill[#3] (#1+0.5, -#2+0.5) circle (0.32);
  }

  \fcell{\xMf}{1}{cMod}
  \fcell{\xWa}{1}{cWf}
  \fcell{\xOc}{1}{cOut}
  \fcell{\xPa}{1}{cScp}
  \fcell{\xAb}{1}{cAcc}

  \fcell{\xMa}{2}{cMod}  \fcell{\xMd}{2}{cMod}
  \fcell{\xWa}{2}{cWf}
  \fcell{\xOa}{2}{cOut}
  \fcell{\xPa}{2}{cScp}  \fcell{\xPc}{2}{cScp}
  \fcell{\xAb}{2}{cAcc}

  \fcell{\xMc}{3}{cMod}
  \fcell{\xWb}{3}{cWf}
  \fcell{\xOc}{3}{cOut}
  \fcell{\xPb}{3}{cScp}
  \fcell{\xAa}{3}{cAcc}

  \fcell{\xMb}{4}{cMod}
  \fcell{\xWa}{4}{cWf}
  \fcell{\xOc}{4}{cOut}
  \fcell{\xPa}{4}{cScp}
  \fcell{\xAb}{4}{cAcc}

  \fcell{\xMc}{5}{cMod}
  \fcell{\xWb}{5}{cWf}
  \fcell{\xOc}{5}{cOut}
  \fcell{\xPa}{5}{cScp}
  \fcell{\xAa}{5}{cAcc}

  \fcell{\xMe}{6}{cMod}
  \fcell{\xWd}{6}{cWf}
  \fcell{\xOc}{6}{cOut}
  \fcell{\xPb}{6}{cScp}  \fcell{\xPc}{6}{cScp}
  \fcell{\xAb}{6}{cAcc}

  \fcell{\xMe}{7}{cMod}
  \fcell{\xWc}{7}{cWf}
  \fcell{\xOd}{7}{cOut}
  \fcell{\xPb}{7}{cScp}
  \fcell{\xAa}{7}{cAcc}

  \fcell{\xMd}{8}{cMod} \fcell{\xMe}{8}{cMod}
  \fcell{\xWc}{8}{cWf}
  \fcell{\xOd}{8}{cOut}
  \fcell{\xPb}{8}{cScp} \fcell{\xPc}{8}{cScp}
  \fcell{\xAa}{8}{cAcc}

  \fcell{\xMe}{9}{cMod}
  \fcell{\xWd}{9}{cWf}
  \fcell{\xOd}{9}{cOut}
  \fcell{\xPb}{9}{cScp} \fcell{\xPc}{9}{cScp}
  \fcell{\xAa}{9}{cAcc}

  \fcell{\xMa}{10}{cMod} \fcell{\xMb}{10}{cMod} \fcell{\xMf}{10}{cMod} \fcell{\xMc}{10}{cMod}
  \fcell{\xMd}{10}{cMod} \fcell{\xMe}{10}{cMod}
  \fcell{\xWa}{10}{cWf} \fcell{\xWb}{10}{cWf} \fcell{\xWc}{10}{cWf} \fcell{\xWd}{10}{cWf}
  \fcell{\xOa}{10}{cOut} \fcell{\xOc}{10}{cOut} \fcell{\xOd}{10}{cOut}
  \fcell{\xPa}{10}{cScp} \fcell{\xPb}{10}{cScp} \fcell{\xPc}{10}{cScp}
  \fcell{\xAa}{10}{cAcc} \fcell{\xAb}{10}{cAcc}

  \newcommand{\countcell}[3]{%
    \node[anchor=center, font=\sffamily\scriptsize#3, text=cSub]
      at (\xV+0.5, -#1+0.5) {#2};
  }
  \countcell{1}{\rev{\countMri}}{}
  \countcell{2}{\countXray}{}
  \countcell{3}{\rev{\countSlide}}{}
  \countcell{4}{\countCt}{}
  \countcell{5}{\countTumor}{}
  \countcell{6}{\countEhrshot}{}
  \countcell{7}{\countDq}{}
  \countcell{8}{\rev{\countCqa}}{}
  \countcell{9}{\rev{\countTte}}{}
  \countcell{10}{\rev{\countTotal}}{\bfseries}

  \foreach \xb in {6.2, 10.6, 14.0, 17.4, 19.8} {
    \draw[gray!50, dashed, line width=0.3pt] (\xb, 0) -- (\xb, -10);
  }
  \draw[gray!60, line width=0.5pt] (-0.05, -9) -- (\xV+1.05, -9);

  \node[anchor=north west, font=\sffamily\scriptsize, text=gray!70,
        text width=10.5cm, align=left]
    at (-1, -10.4) {Filled dot = the task category covers that dimension. Color encodes
      axis. Last column counts the number of tasks in each task category;
      the ALL-tasks row sums across the suite.};

  \end{tikzpicture}
  \caption{\textbf{Coverage of \projectname tasks across five
  evaluation axes.} \rev{Rows are nine task categories.} \rev{Columns} are dimension values within five axes (modality,
  \auditrev{workflow (immediate task purpose)}, output shape, patient/temporal scope, and
  data access regime). \auditrev{Workflow groups overlap in clinical practice: imaging interpretation is diagnostic, tumor slide/area selection assesses specimens, EHR queries/audits manage data, and event/survival model development falls under research.} WSI denotes whole-slide
  (pathology) imaging. A filled dot indicates the
  task category covers that dimension of the axis. The rightmost column reports the number
  of tasks each task category contains. The bottom ``ALL tasks'' row
  \rev{shows the union of observed coverage and reports the total task count}
  (\rev{\countTotal}{}). Most task categories span multiple axes simultaneously
  (e.g., \taskXray is multimodal with both X-ray and
  text inputs, and is both single-patient and longitudinal), giving a
  single \projectname sweep broad evaluative reach.}
  \label{fig:coverage-matrix}
\end{figure}

\section{Cost vs.\ performance trade-off}
\label{appendix:cost-performance}

\begin{figure}[!ht]
  \centering
  \includegraphics[width=0.9\linewidth]{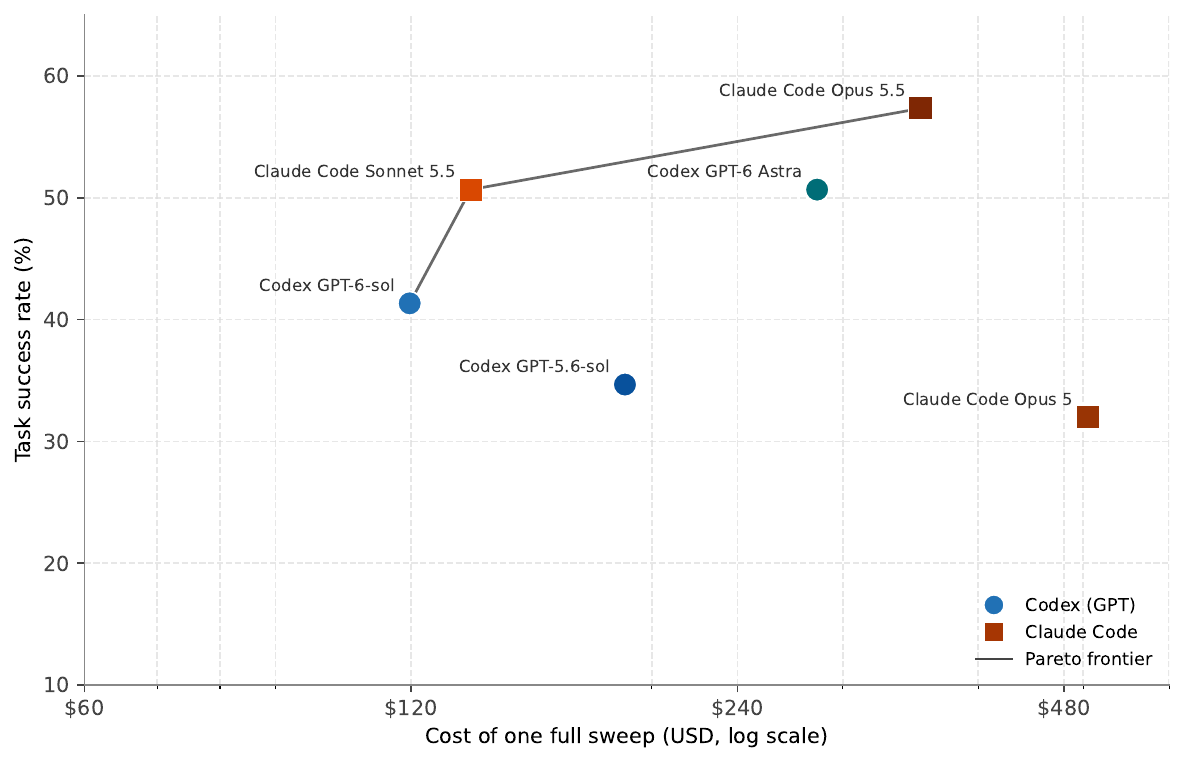}
  \caption{\textbf{Cost vs.\ success rate on \projectname.} Each point is an
  agent; the $x$-axis is the total USD cost of one full sweep over the suite
  of \rev{\countTotal}{} tasks and the $y$-axis is the pooled task success rate
  (trial-weighted, as in Figure~\ref{fig:hero}). \rev{The frontier connects GPT-6-sol, Sonnet-5.5, and Opus-5.5, reflecting the trade-off between lower recorded agent cost and higher task success.}}
  \label{fig:cost-performance}
\end{figure}

Section~\ref{sec:baselines} notes that the best-performing agents are not the
most expensive. Figure~\ref{fig:cost-performance} makes this explicit by
plotting each agent's task success rate against the cost of running it over the
full \projectname{} suite. The $x$-axis is the total USD cost of one full sweep
over the suite on a log scale, and
the $y$-axis is the pooled, trial-weighted task success rate (identical to
Figure~\ref{fig:hero}). \rev{The observed frontier comprises GPT-6-sol, Sonnet-5.5, and Opus-5.5. Sonnet-5.5 achieves 38/75 successes at \$136.15 per sweep, compared with Opus-5.5's 43/75 at \$353.89. Costs are recorded API-equivalent agent estimates and exclude judge calls, infrastructure, and discarded attempts.}

\section{Per-Category Performance Breakdowns}
\label{appendix:per-task}

This appendix collects finer-grained, per-category performance breakdowns that
complement the aggregate results in Section~\ref{sec:baselines}.

\subsection{Per-category results: success rate, time, and cost}
\label{appendix:per-category-results}

Figure~\ref{fig:baseline-heatcircle} summarizes, for every (agent, task
category) cell, three quantities at once---via colour, circle size, and the
\$~tier. Tables~\ref{tab:per-cat-success}--\ref{tab:per-cat-cost} report the
exact values behind that figure so each cell can be read precisely.
Table~\ref{tab:per-cat-success} gives the binary \emph{task success rate} (\%);
Table~\ref{tab:per-cat-time} the mean \emph{wall-clock time} per trial
(minutes), a proxy for effort; and Table~\ref{tab:per-cat-cost} the mean
\emph{cost} per trial (USD). \rev{Every cell uses one attempt per task at xhigh reasoning effort. The six agents use Codex or Claude Code. Complete trial time includes setup and verification; costs are API-equivalent agent estimates. The tables below are generated directly from the same input JSON as the figures.}

\begin{table}[!ht]
\centering\scriptsize
\setlength{\tabcolsep}{3.5pt}
\begin{tabular}{lrrrrrrrrr}
\toprule
Agent & X-ray & CT & MRI & Area & Slides & DQ & CQA & Event & TTE \\
\midrule
Codex GPT-5.6-sol & 60.0 & 10.0 & 0.0 & 70.0 & 30.0 & 25.0 & 0.0 & 66.7 & 60.0 \\
Codex GPT-6 Astra & 60.0 & 30.0 & 20.0 & 80.0 & 30.0 & 37.5 & 33.3 & 100.0 & 100.0 \\
Codex GPT-6-sol & 40.0 & 30.0 & 10.0 & 80.0 & 20.0 & 37.5 & 0.0 & 83.3 & 100.0 \\
Claude Code Opus 5 & 30.0 & 10.0 & 0.0 & 50.0 & 20.0 & 25.0 & 50.0 & 83.3 & 60.0 \\
Claude Code Opus 5.5 & 50.0 & 20.0 & 20.0 & 90.0 & 50.0 & 50.0 & 100.0 & 83.3 & 100.0 \\
Claude Code Sonnet 5.5 & 50.0 & 20.0 & 10.0 & 90.0 & 60.0 & 12.5 & 83.3 & 83.3 & 80.0 \\
\bottomrule
\end{tabular}
\caption{\rev{\textbf{Task success rate (\%)} per agent and task category. Each cell uses one attempt per task. X-ray: report correction; CT: computed tomography abnormality classification; MRI: magnetic resonance imaging question answering; Area/Slides: pathology tumor area/slide selection; DQ: EHR data-quality auditing; CQA: EHR compositional question answering; Event: EHR event modelling; TTE: cancer time-to-event prediction. EHR denotes electronic health records.}}
\label{tab:per-cat-success}
\end{table}

\begin{table}[!ht]
\centering\scriptsize
\setlength{\tabcolsep}{3.5pt}
\begin{tabular}{lrrrrrrrrr}
\toprule
Agent & X-ray & CT & MRI & Area & Slides & DQ & CQA & Event & TTE \\
\midrule
Codex GPT-5.6-sol & 2.2 & 11.4 & 10.4 & 13.4 & 9.8 & 14.8 & 12.4 & 31.2 & 28.3 \\
Codex GPT-6 Astra & 2.2 & 3.9 & 7.5 & 8.8 & 8.4 & 13.2 & 16.4 & 34.7 & 25.2 \\
Codex GPT-6-sol & 2.1 & 5.3 & 8.4 & 11.9 & 6.5 & 17.8 & 15.4 & 20.6 & 17.9 \\
Claude Code Opus 5 & 4.6 & 18.5 & 23.2 & 18.1 & 24.9 & 12.2 & 20.5 & 23.0 & 22.5 \\
Claude Code Opus 5.5 & 1.8 & 8.2 & 33.6 & 11.0 & 22.4 & 10.1 & 11.0 & 20.3 & 12.4 \\
Claude Code Sonnet 5.5 & 1.7 & 5.5 & 19.0 & 8.7 & 9.7 & 9.6 & 17.5 & 31.3 & 22.7 \\
\bottomrule
\end{tabular}
\caption{\rev{\textbf{Mean complete trial wall-clock time (minutes)} per agent and task category. Each cell uses one attempt per task. X-ray: report correction; CT: computed tomography abnormality classification; MRI: magnetic resonance imaging question answering; Area/Slides: pathology tumor area/slide selection; DQ: EHR data-quality auditing; CQA: EHR compositional question answering; Event: EHR event modelling; TTE: cancer time-to-event prediction. EHR denotes electronic health records.}}
\label{tab:per-cat-time}
\end{table}

\begin{table}[!ht]
\centering\scriptsize
\setlength{\tabcolsep}{3.5pt}
\begin{tabular}{lrrrrrrrrr}
\toprule
Agent & X-ray & CT & MRI & Area & Slides & DQ & CQA & Event & TTE \\
\midrule
Codex GPT-5.6-sol & 0.18 & 1.32 & 1.58 & 3.54 & 2.32 & 5.80 & 3.80 & 3.10 & 2.33 \\
Codex GPT-6 Astra & 0.34 & 1.12 & 2.43 & 3.31 & 3.68 & 5.99 & 6.01 & 10.16 & 6.08 \\
Codex GPT-6-sol & 0.14 & 0.47 & 1.21 & 1.05 & 1.24 & 4.84 & 2.92 & 2.18 & 1.87 \\
Claude Code Opus 5 & 1.26 & 6.07 & 13.27 & 5.59 & 14.58 & 4.54 & 5.68 & 1.84 & 3.20 \\
Claude Code Opus 5.5 & 0.26 & 1.44 & 14.57 & 2.43 & 10.27 & 3.87 & 3.79 & 0.92 & 1.00 \\
Claude Code Sonnet 5.5 & 0.13 & 0.60 & 4.14 & 1.23 & 2.55 & 2.24 & 4.10 & 0.58 & 0.73 \\
\bottomrule
\end{tabular}
\caption{\rev{\textbf{Mean API-equivalent agent cost per trial (USD)} per agent and task category. Each cell uses one attempt per task. X-ray: report correction; CT: computed tomography abnormality classification; MRI: magnetic resonance imaging question answering; Area/Slides: pathology tumor area/slide selection; DQ: EHR data-quality auditing; CQA: EHR compositional question answering; Event: EHR event modelling; TTE: cancer time-to-event prediction. EHR denotes electronic health records.}}
\label{tab:per-cat-cost}
\end{table}

\subsection{Task-specific scores}
\label{appendix:per-task:task-specific}

\begin{figure}[t]
  \centering
  \includegraphics[width=\linewidth]{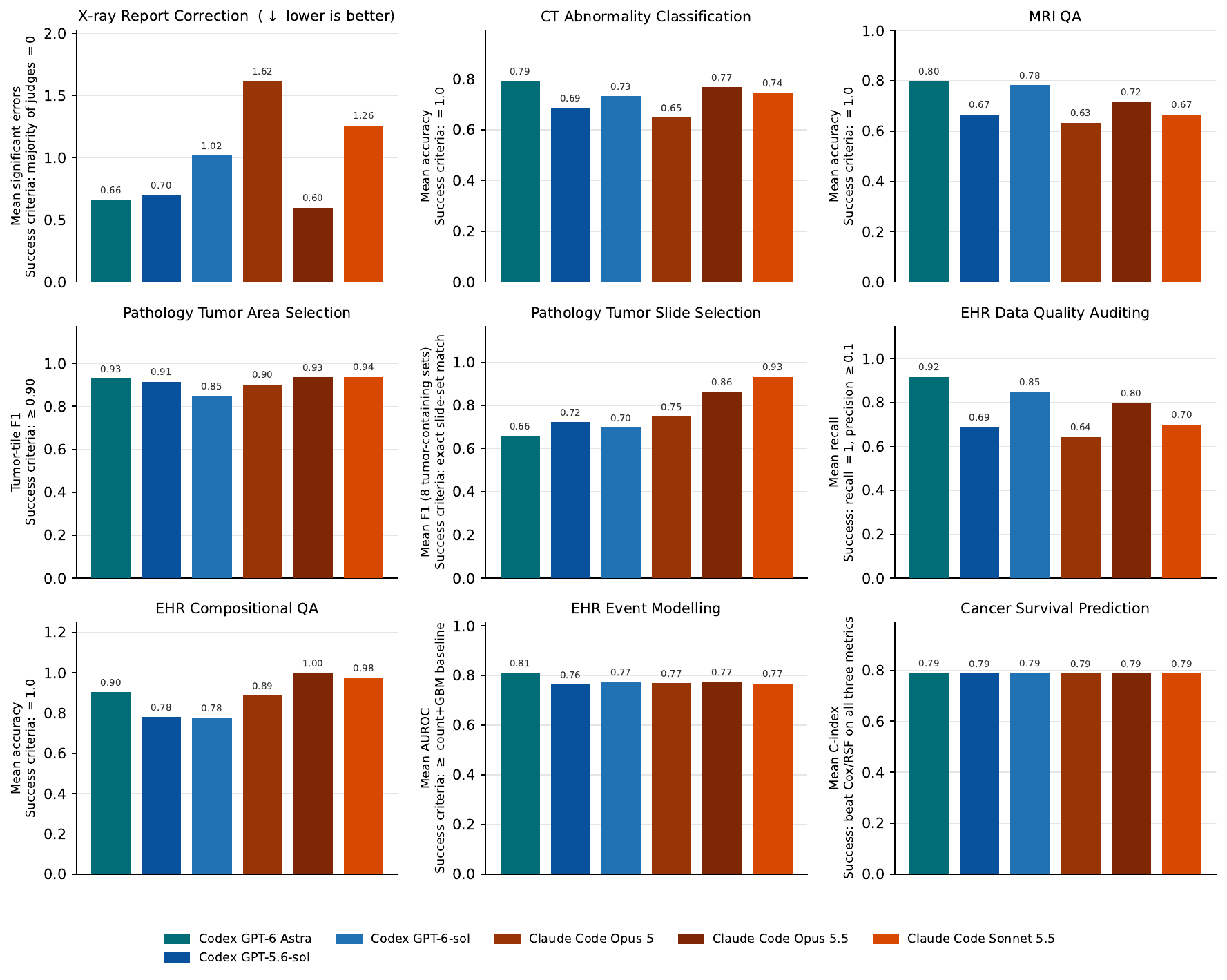}
  \caption{\textbf{Category-specific scores per agent.} One panel per benchmark;
  each bar is an agent. Each panel uses its own $y$-axis because the metrics
  live on different scales. For \taskXray the metric is a mean error count, so
  \emph{lower is better}; all other panels are higher-is-better. \rev{Slide-selection F1 is averaged over the eight tumor-containing sets, excluding tumor-free sets 04 and 10 because the verifier assigns F1 = 0 even for correct empty predictions. Task success rates retain all ten sets. Survival success additionally requires lower censoring-adjusted error and integrated Brier score than both Cox and RSF references; Figure~\ref{fig:survival-tte} reports all three survival metrics.}}
  \label{fig:task-specific}
\end{figure}

The headline task success rate reduces each task to a binary success/failure.
Figure~\ref{fig:task-specific} reports the
\emph{task-specific} metric before converting to binary score
\rev{for all nine categories: mean accuracy for CT, MRI, and compositional QA; tumor-tile F1 for pathology area selection; slide-set F1 for pathology slide selection; cluster recall for data quality; AUROC for event prediction; C-index for cancer survival prediction; and significant-error count for X-ray reports (lower is better). These are means over task instances, not pooled label/question accuracies; The survival C-index panel omits the additional error and calibration gates shown in Figure~\ref{fig:survival-tte}.} Each panel's $y$-axis also notes
the task's success criterion.

\subsection{\rev{\taskCqa: individual question categories versus combined questions}}
\label{appendix:per-task:cqa}

\rev{The CQA task contains six tasks where five individual CQA tasks are an exact disjoint partition of the combined task's 51 questions, patient identifiers, reference answers, and tolerances. All six task packages supply the same 1,020 labeled training examples. We pool question counts, rather than averaging category accuracies. Individual versus combined correct counts are 41/51 versus 37/51 for GPT-5.6-sol, 47/51 versus 50/51 for GPT-6-astra, 40/51 versus 43/51 for GPT-6-sol, 48/51 versus 27/51 for Opus-5, 51/51 in both for Opus-5.5, and 50/51 versus 51/51 for Sonnet-5.5.}

\begin{figure}[t]
  \centering
  \includegraphics[width=\linewidth]{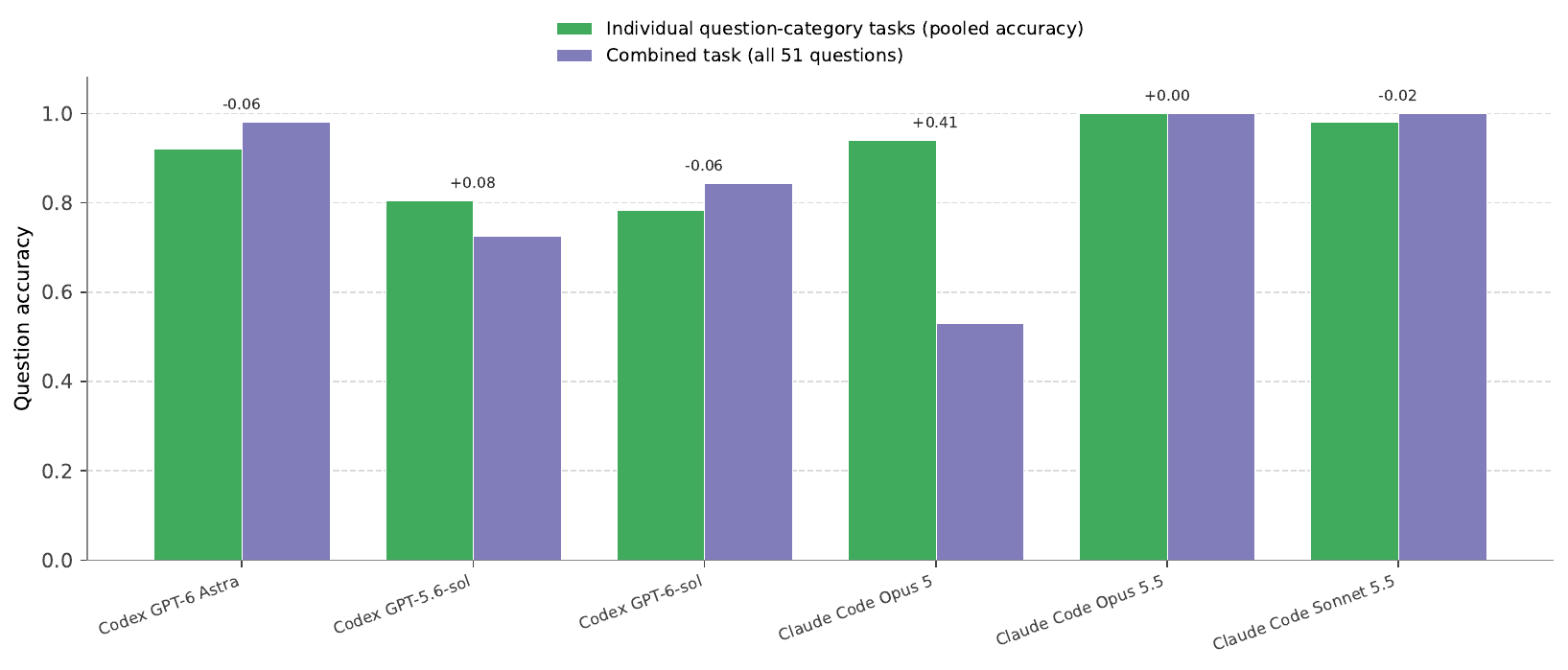}
  \caption{\rev{\textbf{Compositional-QA accuracy: individual categories versus the combined task.} Each bar covers the same 51 questions. Individual accuracy pools correct answers across five category tasks; combined accuracy comes from one all-questions task. The annotated gap is individual minus combined. Each model/task has one attempt}}
  \label{fig:cqa-single-combined}
\end{figure}

\subsection{\rev{\taskEhrshot: AUROC vs.\ computational baselines}}
\label{appendix:per-task:ehr-auroc}

\begin{figure}[t]
  \centering
  \includegraphics[width=\linewidth]{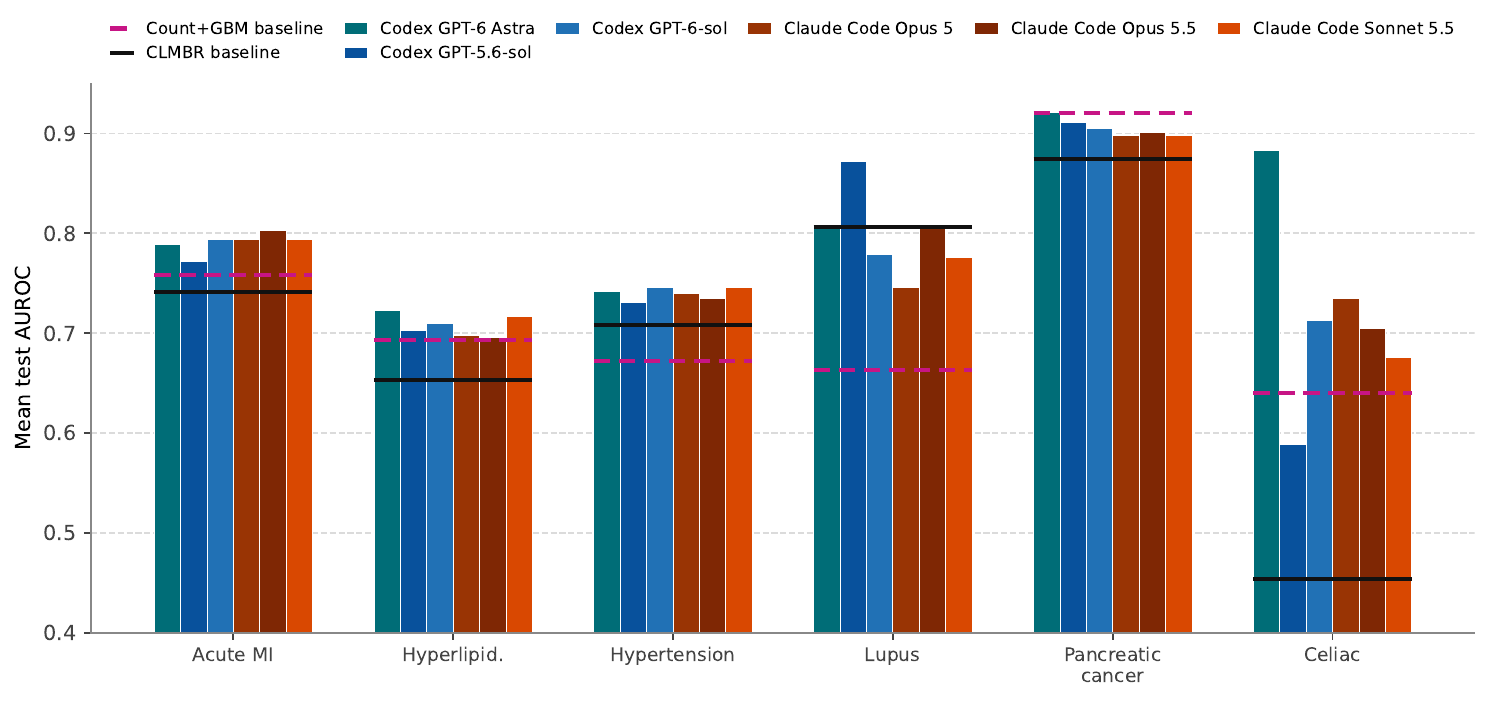}
  \caption{\rev{\textbf{Per-target test AUROC on \taskEhrshot.} Bars show one run per agent and prediction target. The solid black reference is cohort-matched CLMBR; dashed magenta is Count+GBM, the executable pass threshold. Both use the same first-prediction-time test cohort as the agents, rather than the full original EHRSHOT evaluation cohort.}}
  \label{fig:ehr-auroc}
\end{figure}

The \taskEhrshot benchmark asks the agent to predict six
new-onset diagnoses from a longitudinal EHR event log.
\rev{Figure~\ref{fig:ehr-auroc} compares all six agents against two references evaluated on the same first-prediction-time test cohort: CLMBR features with a logistic-regression head and Count+GBM over engineered event counts. Count+GBM defines the binary pass threshold, with zero tolerance margin. GPT-6-astra meets that threshold on all six targets, GPT-5.6-sol on four, and the other agents on five. GPT-5.6-sol meets CLMBR on all six; each other agent meets it on five, with lupus the exception. Our data preparation retains only the first prediction time for each test patient, yielding a subset of the original EHRSHOT test examples. Both references are evaluated on this same subset as the agents, so their AUROC values may differ from those reported in the original EHRSHOT study~\citep{ehrshot}.}

\subsection{\rev{\taskSurvival: comparison with Cox and random survival forests}}
\label{appendix:per-task:survival}
\begin{figure}[p]
  \centering
  \includegraphics[width=0.94\linewidth]{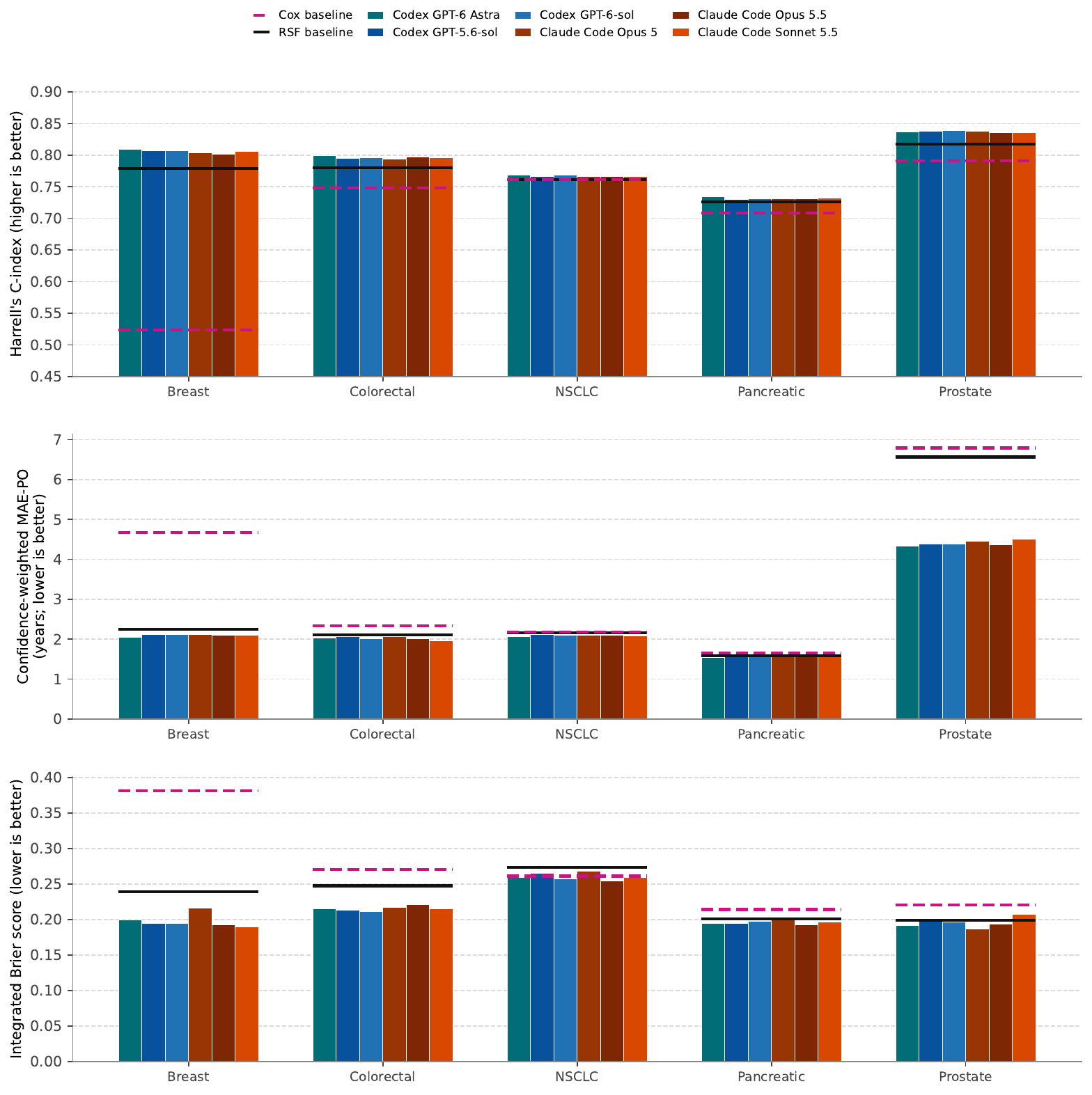}
  \caption{\rev{\textbf{Cancer survival prediction versus cohort-matched Cox/RSF references.} Bars show one attempt per model and cohort; dashed magenta and solid black lines show Cox and RSF, respectively. Higher C-index is better; lower confidence-weighted MAE-PO (years) and IBS are better. Success requires full coverage and strict improvement over both references on all three metrics. IBS evaluates the same scalar-time-to-step-curve representation for agents and references. Official GPT-5.6-sol and GPT-6-sol lung results retain a row-order tie-breaking caveat.}}
  \label{fig:survival-tte}
\end{figure}

\begin{revision}
The five time-to-event survival prediction tasks cover breast, colorectal, non-small-cell lung, pancreatic, and prostate cancer. Agents receive labeled training tables and submit a survival time for every held-out patient. Cox and random survival forest (RSF) references use the same cohort, split, features, and scalar-time representation. Figure~\ref{fig:survival-tte} reports all three scoring metrics rather than only the C-index used in the category summary.

GPT-6-astra, GPT-6-sol, and Opus-5.5 receive 5/5 official passes, Sonnet-5.5 receives 4/5, and GPT-5.6-sol and Opus-5 receive 3/5 each. Every submission has full coverage and a C-index strictly above both references. The censoring-adjusted error gate passes in 29/30 runs, but the integrated Brier score (IBS) gate passes in 25/30. Thus discrimination alone does not explain task success: all five failures miss the IBS threshold (GPT-5.6-sol on lung and prostate, Opus-5 on lung and pancreatic, and Sonnet-5.5 on prostate); Opus-5's pancreatic run also misses the error threshold. Appendix~\ref{appendix:trajectories} contrasts a passing breast-cancer attempt with a failed lung-cancer attempt.
\end{revision}

\section{\projectname Task Categories \label{appendix:benchmarks_task_categories}
}

\rev{The following sections summarize the current nine-category release. Complete versioned instructions, resource limits, and scorer code are in the task files linked from \url{https://github.com/microsoft/HealthAgentBench}. Agent execution permits model APIs only and uses no GPU. The summaries are not verbatim task prompts.}

\subsection{\taskXray: chest X-ray report correction}
\label{appendix:bench:xray_report_correction}

\paragraph{Tags.}
\begin{itemize}
  \item \emph{Modality:} Longitudinal Imaging (chest X-ray) + text.
  \item \auditrev{\emph{Workflow:} Diagnosis.}
  \item \emph{Output shape:} Free-text radiology section (\textsc{Findings}).
  \item \emph{Data access:} PhysioNet credentialed (MIMIC-CXR).
  \item \emph{Number of Tasks} 10
\end{itemize}

\paragraph{Motivation.}
\rev{This task category introduces chest X-ray report correction in \projectname. Accurate chest X-ray reports communicate imaging findings and interval changes that inform patient care.}
The agent
reviews a counterfactual draft \textsc{Findings} for a target chest
X-ray study against the patient's longitudinal imaging history and
produces a corrected \textsc{Findings} that resolves the
clinically-significant errors the draft introduces. Success requires
repeated tool calls to inspect images, integration of visual
observation with textual prior context, and structured-output
discipline. \projectname's contribution is the conversion of a real-world, multimodal
clinical workflow into a verifiable agent task within the data environment of the patients' longitudinal multimodal contexts.

\paragraph{Task setup.}
\begin{itemize}
  \item \emph{Inputs (mounted at \texttt{/data/patient/} with opaque IDs,
    \texttt{study\_NN\_<timestamp>/view\_NN.jpg}):} the patient's prior studies
    (JPGs + free-text reports + timestamps); the target study's non-generative
    sections (\textsc{Examination}, \textsc{Indication}, \textsc{History},
    \textsc{Technique}, \textsc{Comparison}); and the counterfactual draft
    \textsc{Findings} (The draft is a perturbed version of the gold \textsc{Findings} that introduces contradictions by swapping the presence/absence of key findings, the location of findings, or the severity of findings, or the temporal changes without adding any new findings that the gold does not mention. The agent must correct the draft by removing the introduced errors without adding any new findings.)
  \item \emph{\rev{Instruction summary (full contracts are in the released task files):}}
    \begin{quote}\small\itshape
    A draft \texttt{FINDINGS} section for the patient's most recent chest X-ray,
    written by a junior radiologist, is already populated in the target study's
    \texttt{report.txt}; review and correct it. You may edit existing sentences
    but may \emph{not} add statements about findings the draft did not already
    mention, and you submit only the corrected \texttt{FINDINGS} (no
    \textsc{Impression}). Use the chest-X-ray images and the prior studies'
    reports to determine the correct findings. All data under
    \texttt{/data/patient/} belongs to one patient; each subfolder is one study,
    folders sort chronologically, and the highest-numbered \texttt{study\_NN} is
    the target. Set \texttt{final\_answer} in \texttt{/workspace/submission.json}
    to the corrected report, starting with a literal \texttt{FINDINGS:} header.
    You have up to 1 hour.
    \end{quote}
  \item \emph{Required action:} detect and correct the
    clinically-significant errors in the draft (editing existing sentences
    only, adding no new findings).
  \item \emph{Output:} JSON at \texttt{/workspace/submission.json}
    with a \texttt{final\_answer} string that begins with a literal
    \texttt{FINDINGS:} header followed by the corrected \textsc{Findings}
    body (no \textsc{Impression}).
  \item \emph{Hidden (verifier-only):} gold \textsc{Findings} at
    \texttt{/tests/target\_report.txt}; the corpus name and MIMIC subject IDs
    do not appear in any agent-visible path.
  \item \emph{Limits:} agent timeout 3600~s, verifier timeout 900~s,
    2~CPUs, 4~GB RAM, 10~GB storage, \rev{model API access only; no other internet.}
\end{itemize}

\paragraph{Data source.}
\begin{itemize}
  \item \emph{Corpus:} MIMIC-CXR v2.1.0~\citep{PhysioNet-mimic-cxr-2.1.0} + MIMIC-CXR-JPG v2.1.0~\citep{PhysioNet-mimic-cxr-jpg-2.1.0};
    verifier uses CheXprompt~\citep{zambrano2025clinically}.
  \item \emph{Access:} PhysioNet credentialed.
  \item \emph{Version pin:} MIMIC-CXR v2.1.0 PhysioNet release.
\end{itemize}

\paragraph{Tasks in this Category.}
\begin{itemize}
  \item \emph{Count:} 10 cases (\texttt{case\_01..case\_10}).
  \item \emph{Selection criteria:} the 10 cases were manually reviewed by three radiologists to be error-free. The set deliberately mixes longitudinal patients with $\geq$1 prior study (cases 01--05, 07, 09, 10) and single-study patients with no prior (cases 06, 08). Each case's draft FINDINGS is then synthesized at bootstrap by applying one or more documented swap principles (lateralization, severity, comparison-word flip, no-prior/no-change, count, location, negation) to the gold, introducing at least one clinically-significant error the agent must correct.

\end{itemize}

\paragraph{Scoring.}
\begin{itemize}
  \item \emph{Per-trial:} CheXprompt LLM-as-a-judge 5-vote majority on the submitted \textsc{Findings} vs.\ held-out gold.
  \item \emph{Reward (Success Criterion):} 1 iff $\geq 3/5$ votes report zero
    clinically-significant errors.
  \item \emph{Default judge:} \texttt{gpt-5.4}.
  \item \emph{Raw Per-trial Metrics:} mean clinically-significant errors.
  \item \emph{Gold Label Review:} each case's gold \textsc{Findings} were hand-reviewed against the paired chest X-ray and the patient's prior imaging history to ensure they are factually supported and free of clinically-significant errors, so the gold is a reliable target for correction.
\end{itemize}

\paragraph{Anti-cheat \& leak-proofing.}
\begin{itemize}
  \item \emph{Two-service split:} \texttt{bootstrap} holds PhysioNet
    credentials (\texttt{PN\_USER}, \texttt{PN\_PASS}); the
    \texttt{main} agent service has none. Therefore, the agent cannot simply download the gold \textsc{Findings} from PhysioNet.
  \item \emph{ID opacity:} the corpus name, MIMIC subject IDs, and
    study identifiers never appear in agent-visible filenames or
    paths.
  \item gold target \textsc{Findings} are never agent-visible; they live only in the verifier.
\end{itemize}

\paragraph{Bootstrap \& environment.}
\begin{itemize}
  \item \emph{\rev{Tools:}} \rev{Python, NumPy, Pillow, Matplotlib, SciPy, scikit-image, and OpenCV are preinstalled for image processing and rendering; Bash, ripgrep, and jq support file and report inspection.}
  \item \emph{Compose pattern:} two-service. \texttt{bootstrap}
    fetches the credentialed PhysioNet release, extracts each case's
    gold \textsc{Findings} to \texttt{/tests/target\_report.txt},
    enumerates and stages priors into \texttt{/data/patient/}, then
    terminates.
  \item \emph{Gating:} \texttt{main} is gated on bootstrap's clean
    exit via docker-compose's
    \texttt{service\_completed\_successfully} condition.
\end{itemize}

\paragraph{Baseline observations.}
\rev{With one scored attempt per task, category successes are GPT-5.6-sol 6/10; GPT-6-astra 6/10; GPT-6-sol 4/10; Opus-5 3/10; Opus-5.5 5/10; Sonnet-5.5 5/10. Exact costs, complete trial times, and partial metrics are reported in Appendix~\ref{appendix:per-task}.}

\paragraph{Known failure modes.}
\begin{itemize}
  \item Agent introduces a new hallucinated finding while removing one
    of the draft's errors.
  \item Agent leaves a draft error in place because it disagrees with
    a prior study's findings.
  \item Submitted JSON does not parse, or \texttt{final\_answer} is
    missing the required \texttt{FINDINGS:} header.
\end{itemize}

\subsection{\taskCt: chest CT interpretation}
\label{appendix:bench:ct_abnormality}

\paragraph{Tags.}
\begin{itemize}
  \item \emph{Modality:} 3-D imaging (NIfTI volume).
  \item \auditrev{\emph{Workflow:} Diagnosis.}
  \item \emph{Output shape:} Per-volume binary label vector.
  \item \emph{Data access:} HuggingFace gated under OpenRAIL (CT-RATE).
  \item \emph{Number of Tasks} 10
\end{itemize}

\paragraph{Motivation.}
\rev{This task category introduces chest CT abnormality classification in \projectname. Identifying abnormalities on chest CT supports the detection and assessment of thoracic disease.}
Each task in this category requires the agent to inspect a non-contrast chest CT volume and classify it for the presence or absence of a set of clinical findings. The agent must read the volume, interpret the imaging data, and produce a binary label for each requested finding. The tasks test the agent's ability to process volumetric medical imaging data, recognize relevant anatomical structures and pathologies, and make accurate diagnostic classifications based on visual evidence. \projectname's contribution is the conversion of a real-world, 3-D imaging clinical workflow into a verifiable agent task within the data environment of the patients' imaging context. \rev{The agent can explore the volume with the preinstalled analysis stack and write its own local scripts; dependency downloads are blocked during execution.}

\paragraph{Task setup.}
\begin{itemize}
  \item \emph{Inputs (mounted at \texttt{/workspace/data/}):} one
    non-contrast chest CT (NIfTI, 200--700 slices, Hounsfield units) at
    \texttt{/workspace/data/scan.nii.gz}, and a per-volume label list
    (\rev{one label per line, with a case-specific retained label set}) at
    \texttt{/workspace/data/labels.txt}.
  \item \emph{\rev{Instruction summary (full contracts are in the released task files):}}
    \begin{quote}\small\itshape
    You are working inside an environment that contains a single
    non-contrast chest-CT volume and a list of clinical findings to
    evaluate. For each listed finding, decide whether it is present or
    absent in the scan. Write a plain text file at
    \texttt{/workspace/submission/predictions.txt} with one
    \texttt{<label>: yes} or \texttt{<label>: no} line per requested
    label, using the exact name from \texttt{labels.txt}
    (case-insensitive); order does not matter. Solve using only the
    volume on disk; do not look up the dataset's published labels or
    report. The reward is binary --- every label must be correct. You
    have up to 1 hour.
    \end{quote}
  \item \emph{Required action:} inspect the 3-D volume (the canonical
    path is \texttt{nibabel} $\rightarrow$ PNG slices $\rightarrow$
    multimodal view) and classify each requested label.
  \item \emph{Output:} text file at
    \texttt{/workspace/submission/predictions.txt}, one
    \texttt{<label>: yes/no} per line; order is irrelevant (the verifier
    matches by label name).
  \item \emph{Hidden (verifier-only):} gold labels at
    \texttt{tests/gold.json} (gitignored, derived at bootstrap).
  \item \emph{Limits:} agent timeout 3600~s, verifier timeout 300~s,
    2~CPUs, 8~GB RAM, 8~GB storage, \rev{model API access only; no other internet.}
\end{itemize}

\paragraph{Data source.}
\begin{itemize}
  \item \emph{Corpus:} CT-RATE~\cite{ctrate} --- 25\,692 non-contrast
    3-D chest-CT volumes paired with radiology reports; validation
    split of 3\,038 volumes.
  \item \emph{Access:} HuggingFace gated under OpenRAIL;
    \texttt{HF\_TOKEN} required at bootstrap.
\end{itemize}

\paragraph{Trial construction.}
\begin{itemize}
  \item \emph{Count:} 10. \auditrev{Each task contains a volume from CT-RATE}'s validation split.
  \item \emph{Selection:} The 10 tasks are selected to cover disease categories with a range from single disease to multiple diseases.
\end{itemize}

\paragraph{Scoring.}
\begin{itemize}
  \item \emph{Reward (Success Criterion):} all disease labels are predicted correct $\rightarrow$
    1.0, any error $\rightarrow$ 0.0.
  \item \emph{Gold Label Review:} \rev{retained labels are derived from explicit positive/negative phrases in the paired source report; unsupported/conflicting labels are excluded.}
  \item \emph{Raw Per-trial Metrics:} mean accuracy.
\end{itemize}

\paragraph{Anti-cheat \& leak-proofing.}
\begin{itemize}
  \item \emph{Token isolation:} \texttt{bootstrap} holds the
    \texttt{HF\_TOKEN}; \texttt{main} has neither token nor report
    access. Therefore, the agent cannot simply download the gold labels from HuggingFace.
  \item \emph{Gold isolation:} \texttt{tests/gold.json} lives only
    inside the verifier volume.
\end{itemize}

\paragraph{Bootstrap \& environment.}
\begin{itemize}
  \item \emph{\rev{Tools:}} \rev{NiBabel, NumPy, Pillow, Matplotlib, SciPy, scikit-image, and OpenCV are preinstalled for volume loading, slice rendering, and image analysis.}
  \item \emph{Compose pattern:} two-service. \texttt{main} depends on
    \texttt{bootstrap}'s successful completion.
  \item \emph{Build-time data:} bootstrap downloads ~3.5\,GB of NIfTI
    volumes (100--350\,MB each) lazily into \auditrev{a shared host cache} so
    re-runs on the same host avoid re-downloading.
\end{itemize}

\paragraph{Baseline observations.}
\rev{With one scored attempt per task, category successes are GPT-5.6-sol 1/10; GPT-6-astra 3/10; GPT-6-sol 3/10; Opus-5 1/10; Opus-5.5 2/10; Sonnet-5.5 2/10. Exact costs, complete trial times, and partial metrics are reported in Appendix~\ref{appendix:per-task}.}

\paragraph{Known failure modes.}
\begin{itemize}
  \item \emph{Under-calling.} The agent renders appropriate lung and
    mediastinal windows but, not seeing florid disease, defaults subtle
    findings (e.g.\ paraseptal emphysema, small nodules) to ``absent'' ---
    so real findings are missed.
  \item \emph{Over-calling.} \rev{False positives also break exact-match reward. In the current Sonnet sweep, for example, some failed volumes have cardiomegaly false positives alongside missed findings. Reference disagreements are not independently clinically adjudicated here.}
\end{itemize}

\subsection{\taskTumor: pathology whole-slide reasoning}
\label{appendix:bench:tumor_area_selection_pathology}

\paragraph{Tags.}
\begin{itemize}
  \item \emph{Modality:} Pathology whole-slide image (gigapixel).
  \item \auditrev{\emph{Workflow:} Specimen assessment.}
  \item \emph{Output shape:} Tumor-tile coordinate set (JSON).
  \item \emph{Data access:} Public (CAMELYON16).
  \item \emph{Number of Tasks} 10
\end{itemize}

\paragraph{Motivation.}
\rev{This task category introduces tumor-area selection in pathology whole-slide images in \projectname. Localizing tumor tissue in whole-slide images supports biomarker quantification, grading, and tumor-specific morphological analysis.}

The task is challenging for several reasons. First,
whole-slide images are gigapixel in scale, on average around
$100{,}000 \times 100{,}000$ pixels, which makes exhaustive inspection
infeasible and forces multi-resolution reasoning: candidate regions are
located at low magnification and confirmed at high magnification. Second,
tumor is often sparse, or present only as small foci (e.g.,
micro-metastases or isolated tumor-cell clusters) that are easily missed.
Third, the boundary between tumor and normal tissue is frequently ambiguous
on hematoxylin and eosin (H\&E) staining alone, such that even expert
annotations are typically aided by immunohistochemical restaining. These
properties make tumor delineation a non-trivial, decision-intensive task
that requires actively navigating the slide, selectively inspecting local
regions, and integrating evidence under a finite inspection budget. The \projectname contribution is
turning CAMELYON16 into a reproducible Harbor-first agentic task with
one task per slide, hidden gold segmentation, and pooled tile-level
aggregate metrics. We do not require dense pixel-level segmentation; instead, the agent selects the grid tiles it believes contain tumor which is enough in many downstream research tasks. Note that CAMELYON16 comprises  lymph-node whole-slide images from breast-cancer patients, and its reference masks delineate \emph{metastatic} breast carcinoma deposited in the node rather than a primary tumor. Because a metastasis is itself neoplastic tumor tissue, we treat these annotated regions as the tumor area to be selected, and ``tumor-area selection'' here corresponds to localizing nodal tumor (metastatic) deposits.

\paragraph{Task setup.}
\begin{itemize}
  \item \emph{Inputs (mounted into the environment):} one whole-slide
    pathology image at \texttt{/data/slide/current/slide.*} (pyramidal
    \texttt{.tif}, read via OpenSlide/tifffile) and the analysis-grid
    metadata in the public task row \texttt{/workspace/benchmark\_tasks.json}.
  \item \emph{\rev{Instruction summary (full contracts are in the released task files):}}
    \begin{quote}\small\itshape
    You are working in a pathology task environment. The current
    whole-slide image is at \texttt{/data/slide/current/slide.*}, a
    public task row at \texttt{/workspace/benchmark\_tasks.json}, and an
    editable submission at \texttt{/workspace/submission.json}. The
    analysis grid uses \texttt{256x256} tiles at downsample 16, so each
    grid tile spans \texttt{4096x4096} full-resolution (level-0) pixels.
    Identify all and only the tiles that contain tumor and populate
    \texttt{predicted\_tumor\_tiles} with objects of the form
    \texttt{\{"x": <int>, "y": <int>\}} using integer grid
    coordinates; treat non-tissue tiles as non-tumor. You have a budget
    of 1.5 hours; \rev{network access is restricted to the model API.}
    \end{quote}
  \item \emph{Required action:} predict the complete set of tumor tiles
    on the fixed $256\times 256$ grid at down-sample 16; a tile is
    ``tumor'' if $\geq 20\%$ of its area is tumor.
  \item \emph{Output:} JSON at \texttt{/workspace/submission.json}
    with \texttt{predicted\_tumor\_tiles}, a list of
    \texttt{\{x, y\}} grid-coordinate objects. (A
    \texttt{contains\_tumor} boolean is also recorded but is not
    scored here.)
  \item \emph{Hidden (verifier-only):} \rev{precomputed expert-mask-derived tumor-tile keys in the separate verifier, never mounted into the agent service.}
  \item \emph{Limits:} agent timeout 5400~s, verifier timeout 3600~s,
    2~CPUs, 8~GB RAM, 20~GB storage, \rev{model API access only; no other internet.}
\end{itemize}

\paragraph{Data source.}
\begin{itemize}
  \item \emph{Corpus:} CAMELYON16~\rev{\citep{camelyon16}} (AWS S3 public mirror at
    \texttt{s3://camelyon-dataset/CAMELYON16/}); slides are pyramidal
    \texttt{.tif} read via OpenSlide.
  \item \emph{Access:} public, no credentials.
\end{itemize}

\paragraph{Tasks in this Category.}
\begin{itemize}
  \item \emph{Count:} \auditrev{10 tasks, each containing a CAMELYON16 tumor slide}
  \item \emph{Selection:} The slides inherit a range of tumor sizes from CAMELYON16 (gold regions span $\approx$20--160 tiles) and varied metastasis morphology; smaller deposits are missed even by strong agents.
\end{itemize}

\paragraph{Scoring.}
\begin{itemize}
  \item \emph{Reward (Success Criterion):} Calculate tile-level F1 of the predicted tile set
    against the hidden gold mask; a trial passes (reward 1.0) iff
    tile-F1 $\geq 0.90$, else 0.0. We choose 0.9 F1 rather than 1.0 F1 to allow for some small differences due to ambiguity in the tumor boundary.
  \item \emph{Raw Per-trial Metrics:} tile-level F1.
\end{itemize}

\paragraph{Anti-cheat \& leak-proofing.}
\begin{itemize}
  \item \emph{Gold isolation:} \rev{expert-mask-derived tile keys live only in the separate verifier.}
  \item \emph{Credential surface:} none (datasets are public).
\end{itemize}

\paragraph{Bootstrap \& environment.}
\begin{itemize}
  \item \emph{\rev{Tools:}} \rev{OpenSlide, tifffile, imagecodecs, NumPy, Pillow, and Matplotlib are preinstalled for whole-slide access and rendering, alongside SciPy, scikit-image, OpenCV, and scikit-learn for local analysis. Agents write their own analysis scripts.}
  \item \emph{Compose pattern:} two-service (one-shot \texttt{bootstrap}
    then \texttt{main}), with lazy public-data download into
    \auditrev{a shared host cache}.
  \item \emph{Cache sharing:} multiple containers on the same host
    share the cache so the gigapixel slides download once per host,
    not once per trial.
\end{itemize}

\paragraph{Baseline observations.}
\rev{With one scored attempt per task, category successes are GPT-5.6-sol 7/10; GPT-6-astra 8/10; GPT-6-sol 8/10; Opus-5 5/10; Opus-5.5 9/10; Sonnet-5.5 9/10. Exact costs, complete trial times, and partial metrics are reported in Appendix~\ref{appendix:per-task}.}

\paragraph{Known failure modes.}
\begin{itemize}
  \item Agent skims via thumbnail and misses small tumor foci.
  \item Agent over-segments tissue artifacts as tumor.
  \item Agent confuses adjacent benign tissue (e.g.\ inflammation)
    with tumor.
\end{itemize}

\subsection{\texttt{ehrshot}: longitudinal clinical event prediction}
\label{appendix:bench:ehrshot}

\paragraph{Tags.}
\begin{itemize}
  \item \emph{Modality:} Longitudinal EHR timeline (tabular events).
  \item \auditrev{\emph{Workflow:} Research.}
  \item \emph{Output shape:} Per-patient probability (CSV).
  \item \emph{Data access:} Redivis-gated (Stanford STARR via EHRSHOT).
  \item \emph{Number of Tasks:} 6
\end{itemize}

\paragraph{Motivation.}
\rev{This task category introduces longitudinal clinical event prediction from EHRs in \projectname. Predicting future clinical events from longitudinal EHRs supports risk stratification and the identification of patients who may need closer follow-up.}
The agent must learn a prediction rule from labelled training data
and apply it to an unlabelled held-out test set. The
\projectname contribution is packaging the EHRSHOT raw-timeline
benchmark as a reproducible Harbor-first task family with one-click
container-side data download, hidden test labels enforced by
docker-compose volume topology, mechanical leak-proofing on the
event slice, and pass-the-baseline success/failure gates.

\paragraph{Task setup.}
\begin{itemize}
  \item \emph{Inputs (mounted at \texttt{/data/}):} training labels
    (\texttt{train\_labels.csv}), validation labels
    (\texttt{val\_labels.csv}), an unlabelled test set
    (\texttt{test\_examples.csv}, no labels), and a per-task leak-proof
    slice of the longitudinal event log (\texttt{events.csv}; for test
    patients only events with ${\rm start} < T_{\rm first}$ are kept). Notice that the test set differs from the original EHRSHOT test set: we take the original test set and only retain the first prediction-time event slice for each test patient, so the agent must predict from partial timelines without any future information.
  \item \emph{\rev{Instruction summary (hyperlipidemia target):}}
    \begin{quote}\small\itshape
    Predict whether the patient will have their first diagnosis of
    hyperlipidemia (code \texttt{SNOMED/55822004} and its ontology
    children) within the next year. Predictions are made at 11:59 PM on
    the day of discharge from an inpatient visit; discharges where the
    patient already has the diagnosis are ignored. Learn a strategy from
    the labelled \texttt{train} and \texttt{val} splits and apply it to a
    held-out \texttt{test} split given without labels, using only events
    with \texttt{start} before each prediction time. Write
    \texttt{/workspace/submission/predictions.csv} with columns
    \texttt{patient\_id,prediction\_time,probability} (continuous in
    $[0,1]$). Scored by AUROC. \rev{You have one hour and model API access only.}
    \end{quote}
  \item \emph{Required action:} learn a prediction strategy from
    train + val, apply to the test set.
  \item \emph{Output:} per-patient probabilities at
    \texttt{/workspace/submission/predictions.csv}, columns
    \texttt{patient\_id,prediction\_time,probability}; scored by AUROC.
  \item \emph{Hidden (verifier-only):} test labels live on a
    \texttt{/tests/} mount that is \emph{not} mounted into \texttt{main};
    the Redivis API token never reaches \texttt{main}.
  \item \emph{Limits:} agent timeout 3600~s (1-hour budget), verifier
    timeout 300~s, 16~CPUs, 64~GB RAM, \rev{64~GiB disk; model API access only inside the agent service.}
\end{itemize}

\paragraph{Data source.}
\begin{itemize}
  \item \emph{Corpus:} EHRSHOT~\cite{ehrshot} on the Stanford STARR
    EHR cohort (6\,739 patients, 41.6M observations).
  \item \emph{Access:} Redivis-gated; access via
    \texttt{\textasciitilde/.redivis/api\_token}.
\end{itemize}

\paragraph{Tasks in this Category.}
\begin{itemize}
  \item \emph{Count:} 6 tasks
  \item \emph{Selection Criteria} We select 6 tasks from EHRSHOT benchmark that focus on new-onset diagnosis predictions
    (hyperlipidemia, celiac disease, acute myocardial infarction,
    pancreatic cancer, systemic lupus (SLE), essential hypertension).
\end{itemize}

\paragraph{Scoring.}
\begin{itemize}
  \item \emph{Reward (Success Criterion):} AUROC is computed on agent's submission against the held-out test labels. 1 iff AUROC $\geq$ the task's count+LightGBM
    baseline (per-task baseline JSON hardcoded in the generator),
    else 0.
  \item \emph{Raw Per-trial Metrics:} test AUROC.
\end{itemize}

\paragraph{Anti-cheat \& leak-proofing.}
\begin{itemize}
  \item \emph{Event-slice integrity:} for test patients, events are
    kept only where \texttt{start < T\_first} (the patient's first
    prediction time) and any future \texttt{end} value is blanked;
    train/val patients keep their full timelines.
  \item \emph{Test-label isolation:} test labels are never committed
    to git; bootstrap writes them to a \texttt{/tests/} mount that
    \texttt{main} does not see.
\end{itemize}

\paragraph{Bootstrap \& environment.}
\begin{itemize}
  \item \emph{\rev{Tools:}} \rev{pandas, NumPy, PyArrow, and Polars support table processing; SciPy, scikit-learn, LightGBM, and XGBoost support feature engineering and model fitting. Matplotlib is also preinstalled.}
  \item \emph{Compose pattern:} two-service. \texttt{bootstrap} reads
    \texttt{\textasciitilde/.redivis/api\_token}, downloads the
    ~4\,GB bundle, slices the event log, partitions splits, writes
    test labels to \texttt{/tests/}, and terminates.
  \item \emph{Resource profile:} 16~CPUs, 64~GB RAM, \rev{64~GiB disk; no GPU.}
\end{itemize}

\paragraph{Baseline observations.}
\rev{With one scored attempt per task, category successes are GPT-5.6-sol 4/6; GPT-6-astra 6/6; GPT-6-sol 5/6; Opus-5 5/6; Opus-5.5 5/6; Sonnet-5.5 5/6. Exact costs, complete trial times, and partial metrics are reported in Appendix~\ref{appendix:per-task}.}

\paragraph{Known failure modes.}
\begin{revision}
\begin{itemize}
  \item Validation tuning can fail to generalize to the held-out cohort; Sonnet-5.5's pancreatic-cancer AUROC is 0.8977, below the 0.9204 reference despite valid complete predictions.
  \item Uncensored visit ends can contaminate training/validation predictors even when starts precede prediction time. Test future ends are blanked by bootstrap; audits found this timing caveat in acute-MI code, not demonstrated hidden-test label access.
  \item Additional labels reconstructed from labeled-cohort future outcomes must be distinguished from predictor features. Such training augmentation is not hidden-test leakage when test patients are excluded.
\end{itemize}
\end{revision}

\subsection{\taskDq: EHR data-quality auditing}
\label{appendix:bench:ehr_data_quality}

\paragraph{Tags.}
\begin{itemize}
  \item \emph{Modality:} Tabular EHR.
  \item \auditrev{\emph{Workflow:} Data management.}
  \item \emph{Output shape:} Flagged-row CSV (\texttt{table,\_row\_id}).
  \item \emph{Data access:} Public (MIMIC-IV demo + synthetic errors).
  \item \emph{Number of Tasks:} 8
\end{itemize}

\paragraph{Motivation.}
\rev{This task category introduces EHR data-quality auditing in \projectname. Detecting errors in EHR data helps prevent unreliable records from undermining clinical analyses and predictive models.}
The agent
acts as an auditor of a structured EHR dataset, flagging rows that
violate clinical plausibility. The \projectname contribution is a
deterministic, reproducible synthetic-error benchmark with hidden
labels, source-name obfuscation, and scoring that combines
cluster-level recall with row-level precision.

\paragraph{Task setup.}
\begin{itemize}
  \item \emph{Inputs (mounted at \texttt{/workspace/data/csv/}):}
    the corrupted EHR-demo subset as gzipped CSVs
    (\texttt{<table>.csv.gz}), spanning 8 hospital and ICU tables. The corruptions are generated by a deterministic synthetic-error recipe seeded at Docker build time including the following error types: impossible values (range-extreme, decimal-shift, unit-confusion, and unit-label mismatch), inconsistencies (conflicting/duplicate records both in-table and cross-table), and demographic contradictions (patient demographics that contradict other records).
  \item \emph{\rev{Instruction summary (demographic-conflict task):}}
    \begin{quote}\small\itshape
    You are working inside a task environment that contains a copy of an
    EHR dataset under \texttt{/workspace/data/}. Do a data-quality check
    and flag data-entry errors belonging to certain error categories.
    Flag all errors in the category \emph{demographic contradictions} ---
    the patient's recorded demographic information contradicts other
    evidence about that patient. Submit a CSV at
    \path{/workspace/submission/flagged_rows.csv} with columns
    \texttt{table} and \texttt{\_row\_id}. You have up to 1 hour; do not
    look up solutions on the internet.
    \end{quote}
  \item \emph{Instruction disclosure:} \rev{the standard instruction names the target error family/families; the paired \texttt{\_hard} variant requests a generic audit without that taxonomy.}
  \item \emph{Required action:} flag rows the agent believes are
    problematic.
  \item \emph{Output:} CSV with the schema \texttt{table,\_row\_id}
    at \path{/workspace/submission/flagged_rows.csv}.
  \item \emph{Limits:} agent timeout 3600~s, verifier timeout 600~s,
    2~CPUs, 4~GB RAM, 10~GB storage, \rev{model API access only; no other internet.}
\end{itemize}

\paragraph{Data source.}
\begin{itemize}
  \item \emph{Corpus:} MIMIC-IV demo (v2.2)~\citep{PhysioNet-mimic-iv-demo-2.2} with a deterministic
    synthetic-error injection recipe applied at Docker build time.
  \item \emph{Pin:} the synthetic-error recipe is seeded; gold is
    derived alongside the corruption.
\end{itemize}

\paragraph{Tasks in this Category.}
\begin{itemize}
  \item \emph{Count:} 8 tasks
  \item \emph{Task Names}: \rev{four standard tasks---demographic conflict, impossible value, inconsistency, and combined---plus a paired \texttt{\_hard} variant of each. Standard/hard pairs share identical data, labels, and scoring, differing in category guidance.}

\end{itemize}

\paragraph{Scoring.}
\begin{itemize}
  \item \emph{Reward (Success Criterion)} a trial passes iff cluster-level
    recall is $1.0$ (every injected error cluster has at least one
    flagged member). We choose recall as \auditrev{the source database may already contain errors}. To guard against the agent flooding the submission with false positives, we also require that precision is at least \rev{$0.1$;}

  \item \emph{Raw Per-trial Metrics:} mean cluster-level recall. A cluster is a set of rows that are all part of the single injected error (e.g., a set of conflicting records). The mean cluster-level recall is computed as the fraction of clusters for which at least one row was flagged by the agent.
  \item \emph{Gold:} \auditrev{The seeded error-injection recipe defines required clusters (7 demographic, 30 impossible-value, 32 inconsistency, or 69 combined). Additional reviewed source-error rows can receive precision credit but do not enter the required-recall denominator.}
\end{itemize}

\paragraph{Anti-cheat \& leak-proofing.}
\begin{itemize}
  \item \emph{\rev{Runtime egress control:}} \rev{the agent phase permits model APIs only, preventing retrieval of clean upstream tables during the run. Bootstrap/build downloads are separate from agent execution.}

\end{itemize}

\paragraph{Bootstrap \& environment.}
\begin{itemize}
  \item \emph{\rev{Tools:}} \rev{Python, pandas, NumPy, DuckDB, PyArrow, and Polars are preinstalled for table inspection and cross-table queries; Bash, ripgrep, and jq support command-line exploration.}
  \item \emph{Compose pattern:} \rev{an agent service with a separate offline verifier.}
  \item \emph{Build-time work:} demo CSVs are downloaded and
    corrupted deterministically at build time using a seeded
    synthetic-error recipe; the gold set is computed alongside and
    held by the verifier.
\end{itemize}

\paragraph{Baseline observations.}
\rev{With one scored attempt per task, category successes are GPT-5.6-sol 2/8; GPT-6-astra 3/8; GPT-6-sol 3/8; Opus-5 2/8; Opus-5.5 4/8; Sonnet-5.5 1/8. Exact costs, complete trial times, and partial metrics are reported in Appendix~\ref{appendix:per-task}.}

\paragraph{Known failure modes.}
\begin{itemize}
  \item Agent under-flags: misses subtle conflicting/duplicate records.
\end{itemize}

\subsection{\rev{\taskMri: knee MRI interpretation}}
\label{appendix:bench:mri}
\begin{revision}
\paragraph{Tags.}
\begin{itemize}
  \item \emph{Modality:} 3-D knee MRI (NumPy arrays).
  \item \auditrev{\emph{Workflow:} Diagnosis.}
  \item \emph{Output shape:} Six binary answers across three examinations.
  \item \emph{Data access:} Credentialed Redivis access to MRNet.
  \item \emph{Number of Tasks:} 10.
\end{itemize}

\paragraph{Motivation.}
This task category introduces knee MRI interpretation for anterior cruciate ligament (ACL) and meniscal tears in \projectname. Identifying ACL and meniscal tears on knee MRI supports the assessment of knee injuries and treatment planning.
The agent must navigate volumetric knee images, select informative views, and identify ACL and meniscal tears. \projectname packages multiple examinations into one terminal-based task, requiring agents to render the supplied arrays and integrate evidence across slices and imaging planes.

\paragraph{Task setup.}
\begin{itemize}
  \item \emph{Inputs:} three independent patients' examinations under \path{/workspace/mri/exam_1/} through \path{/workspace/mri/exam_3/}. Each contains \texttt{axial.npy}, \texttt{coronal.npy}, and \texttt{sagittal.npy}: uint8 arrays of shape slices $\times256\times256$.
  \item \emph{Instruction summary:}
    \begin{quote}\small\itshape
    Review each examination and determine whether an ACL tear and a meniscal tear are present. Render the views you need from the arrays. Answer each examination from its own images and submit all six binary answers. You have up to two hours.
    \end{quote}
  \item \emph{Required action:} inspect the images and classify both findings for each patient. ACL questions include partial and complete tears; meniscal questions specify surface-reaching signal on at least two slices or meniscal deformity, with exclusions stated in the task prompt.
  \item \emph{Output:} \path{/workspace/submission.json}, with keys \texttt{exam\_1}, \texttt{exam\_2}, and \texttt{exam\_3}, each containing integer-valued \texttt{acl} and \texttt{meniscus} fields (0 or 1).
  \item \emph{Limits:} agent timeout 7200~s, verifier timeout 120~s, 2~CPUs, 4~GiB RAM, 10~GiB storage, no GPU, and model API access only.
\end{itemize}

\paragraph{Data source.}
\begin{itemize}
  \item \emph{Corpus:} MRNet knee MRI examinations and released validation labels~\citep{bien2018mrnet}, staged through Redivis.
  \item \emph{Source verification:} manifest checksums and normalized examination-ID joins link arrays to labels. The source-table integrity audit excludes an unmatched examination; it is not independent radiologist adjudication.
\end{itemize}

\paragraph{Tasks in this Category.}
\begin{itemize}
  \item \emph{Count:} ten environments, each bundling three examinations and six questions.
  \item \emph{Construction:} a task-specific manifest fixes the source examinations and series; opaque examination names replace source identifiers in the agent workspace.
\end{itemize}

\paragraph{Scoring.}
\begin{itemize}
  \item \emph{Reward (Success Criterion):} a valid submission must match all six reference answers; any incorrect answer yields reward 0.
  \item \emph{Raw Per-trial Metrics:} answer accuracy.
  \item \emph{Gold:} released validation labels joined to the staged examinations during bootstrap.
\end{itemize}

\paragraph{Anti-cheat \& leak-proofing.}
\begin{itemize}
  \item \emph{Isolation:} labels, source manifests, the download cache, and Redivis credentials remain outside the agent service; the staged arrays are mounted read-only.
\end{itemize}

\paragraph{Bootstrap \& environment.}
\begin{itemize}
  \item \emph{Data staging:} a separate bootstrap service downloads missing arrays and label tables, verifies source checksums, and stages images and verifier-only labels.
  \item \emph{Tools:} NumPy, Pillow, Matplotlib, SciPy, scikit-image, and OpenCV are preinstalled for local image processing and rendering; verification runs offline in a separate container.
\end{itemize}

\paragraph{Baseline observations.}
With one attempt per task, GPT-6-astra and Opus-5.5 pass 2/10; GPT-6-sol and Sonnet-5.5 pass 1/10; GPT-5.6-sol and Opus-5 pass none.

\paragraph{Known failure modes.}
\begin{itemize}
  \item \emph{Image interpretation:} agents can misclassify ACL or meniscal findings despite inspecting multiple planes and revisiting uncertain cases. For example, Opus-5.5 set~03 disagrees with the meniscus reference for one examination; that single mismatch fails the six-answer gate.
  \item \emph{Image-context handling:} in Sonnet-5.5 set~05, earlier images are removed from context and the agent acknowledges unsupported visual descriptions. It subsequently reopens images, but the contribution of context loss to the final errors is unclear.
\end{itemize}
\end{revision}

\subsection{\rev{\taskSlide: identifying tumor-containing slides}}
\label{appendix:bench:slides}
\begin{revision}
\paragraph{Tags.}
\begin{itemize}
  \item \emph{Modality:} Whole-slide H\&E pathology images.
  \item \auditrev{\emph{Workflow:} Specimen assessment.}
  \item \emph{Output shape:} Set of tumor-containing slide identifiers.
  \item \emph{Data access:} Public NCI Genomic Data Commons (GDC).
  \item \emph{Number of Tasks:} 10.
\end{itemize}

\paragraph{Motivation.}
This task category introduces tumor-bearing slide selection from collections of pathology images in \projectname. Identifying tumor-bearing pathology slides enables selection of appropriate tissue for downstream cancer analyses.
The agent must review a collection of large pathology images and identify every slide containing malignant cells. Unlike tumor-area selection, this task requires slide-level decisions across eight images, including sets with no tumor slides. The agent chooses how to navigate image resolutions and sample candidate regions.

\paragraph{Task setup.}
\begin{itemize}
  \item \emph{Inputs:} eight opaque whole-slide images, \texttt{slide\_1.svs} through \texttt{slide\_8.svs}, under \path{/data/slides/}.
  \item \emph{Instruction summary:}
    \begin{quote}\small\itshape
    Identify every slide containing malignant tumor cells, whether primary or metastatic. Any number of the eight slides, including none, may contain tumor. Submit the list of tumor-containing slides. You have up to two hours.
    \end{quote}
  \item \emph{Required action:} inspect whole-slide overviews and regions as needed, and decide which slides contain tumor.
  \item \emph{Output:} \path{/workspace/submission.json}, with a \texttt{tumor\_slides} list of slide filenames; an empty list represents no tumor slides.
  \item \emph{Limits:} agent timeout 7200~s, verifier timeout 600~s, 2~CPUs, 8~GiB RAM, 20~GiB storage, no GPU, and model API access only.
\end{itemize}

\paragraph{Data source.}
\begin{itemize}
  \item \emph{Corpus:} public TCGA frozen top/bottom tissue-section images from GDC~\citep{grossman2016gdc}; these are not FFPE diagnostic slides.
  \item \emph{Labels:} GDC documents tumor/normal tissue-type metadata~\citep{gdcslides}. Our task includes Primary Tumor or Metastatic samples only when their recorded pathologist-reviewed \texttt{percent\_tumor\_cells} exceeds zero, and Solid Tissue Normal samples only when that percentage equals zero. Slides without a recorded review are excluded.
\end{itemize}

\paragraph{Tasks in this Category.}
\begin{itemize}
  \item \emph{Count:} ten sets of eight slides.
  \item \emph{Construction:} fixed source-slide mappings define each set; the agent sees opaque filenames and is not told how many slides contain tumor.
\end{itemize}

\paragraph{Scoring.}
\begin{itemize}
  \item \emph{Reward (Success Criterion):} the submitted slide set must exactly match the reference tumor-slide set.
  \item \emph{Raw Per-trial Metrics:} slide-level precision, recall, and F1.
  \item \emph{Gold:} hidden slide labels based on the source sample types and recorded pathology reviews.
\end{itemize}

\paragraph{Anti-cheat \& leak-proofing.}
\begin{itemize}
  \item \emph{Isolation:} source mappings and download URLs are available only to bootstrap; hidden labels remain in the separate verifier.
\end{itemize}

\paragraph{Bootstrap \& environment.}
\begin{itemize}
  \item \emph{Data staging:} a one-shot bootstrap service downloads public slides into \auditrev{a shared host cache} and stages the eight selected images for the agent.
  \item \emph{Tools:} OpenSlide, tifffile, imagecodecs, and general Python image-analysis packages are preinstalled. The verifier runs offline in a separate container.
\end{itemize}

\paragraph{Baseline observations.}
With one attempt per task, Sonnet-5.5 passes 6/10 and Opus-5.5 passes 5/10; GPT-5.6-sol and GPT-6-astra pass 3/10, while GPT-6-sol and Opus-5 pass 2/10 (Appendix~\ref{appendix:per-task}).

\paragraph{Known failure modes.}
\begin{itemize}
  \item Missing a tumor slide or including a normal slide breaks exact-set correctness.
  \item Reviewed Sonnet runs show actual slide inspection, but failed sets~02, 03, and~10 also record removed image history and acknowledged unsupported narration.
\end{itemize}
\end{revision}

\subsection{\rev{\taskCqa: raw-EHR clinical calculations}}
\label{appendix:bench:cqa}
\begin{revision}
\paragraph{Tags.}
\begin{itemize}
  \item \emph{Modality:} Structured EHR tables and clinical questions.
  \item \auditrev{\emph{Workflow:} Data management.}
  \item \emph{Output shape:} Patient-specific numerical answers or \texttt{not\_applicable}.
  \item \emph{Data access:} Public MIMIC-IV demo v2.2.
  \item \emph{Number of Tasks:} 6.
\end{itemize}

\paragraph{Motivation.}
This task category introduces compositional clinical question answering over raw EHRs in \projectname. Accurate clinical calculations from longitudinal EHRs support assessment of disease severity and patient status.
We adapt CodeClinic's compositional information-seeking task~\citep{ossowski2026codeclinic} into terminal environments that require agents to reconstruct clinical calculations from raw hospital and ICU records. Agents must connect patient, admission, and ICU-stay identifiers, implement clinical concepts, and validate dataset-specific conventions against supplied training examples.

\paragraph{Task setup.}
\begin{itemize}
  \item \emph{Inputs:} \path{/workspace/data/mimic4.db}, a DuckDB database containing raw \texttt{mimiciv\_hosp} and \texttt{mimiciv\_icu} tables; patient-linked questions in \path{/workspace/questions.json}; and 1,020 labeled examples in \path{/workspace/train.json}. No precomputed derived tables are supplied.
  \item \emph{Instruction summary:}
    \begin{quote}\small\itshape
    Answer every clinical question by querying the supplied records. Validate calculations against the labeled training examples, which define dataset-specific item IDs, time windows, units, and edge cases. Fill each answer in the supplied submission template. You have up to one hour.
    \end{quote}
  \item \emph{Required action:} retrieve the relevant records, compute the requested quantities, and check calculation conventions against training examples.
  \item \emph{Output:} a JSON list in \path{/workspace/submission.json}, with exactly one \texttt{question\_id}/\texttt{answer} pair per question. Answers are numbers or the literal \texttt{not\_applicable}; unfilled \texttt{null} answers score as incorrect.
  \item \emph{Limits:} agent timeout 3600~s, verifier timeout 300~s, 2~CPUs, 4~GiB RAM, 8~GiB storage, no GPU, and model API access only.
\end{itemize}

\paragraph{Data source.}
\begin{itemize}
  \item \emph{Corpus:} MIMIC-IV Clinical Database Demo v2.2~\citep{PhysioNet-mimic-iv-demo-2.2}, with clinical question variants adapted from CodeClinic.
  \item \emph{Training:} twenty labeled examples per variant, shared across the six environments. Training and test-question patients are disjoint.
\end{itemize}

\paragraph{Tasks in this Category.}
\begin{itemize}
  \item \emph{Count:} five category tasks partition 51 questions into direct (15), aggregation (8), comparative (7), derived (7), and temporal (14) types. A sixth task contains all 51 questions.
  \item \emph{Coverage:} twelve clinical concepts span 51 question variants. The individual and combined environments reuse the same questions; they do not constitute 102 independent clinical questions.
\end{itemize}

\paragraph{Scoring.}
\begin{itemize}
  \item \emph{Reward (Success Criterion):} all questions must be answered correctly in a valid submission. Integer answers require absolute error $<0.5$; continuous answers must satisfy their stated relative/absolute tolerance. \texttt{not\_applicable} must match the reference exactly.
  \item \emph{Raw Per-trial Metrics:} answer accuracy and per-concept accuracy.
  \item \emph{Reference conventions:} executable calculations define the answers. These checks establish reference agreement rather than independent clinical review.
\end{itemize}

\paragraph{Anti-cheat \& leak-proofing.}
\begin{itemize}
  \item \emph{Isolation:} the agent mounts only the raw database read-only; bootstrap files, its cache, and test-answer keys remain outside the agent container. The verifier receives the submitted answers in a separate offline container.
\end{itemize}

\paragraph{Bootstrap \& environment.}
\begin{itemize}
  \item \emph{Data staging:} bootstrap downloads the public demo on a cache miss and builds the raw-only DuckDB database before the agent starts.
  \item \emph{Tools:} Bash, Python, DuckDB, pandas, NumPy, and PyArrow are preinstalled.
\end{itemize}

\paragraph{Baseline observations.}
With one attempt per task, Opus-5.5 passes 6/6, Sonnet-5.5 5/6, Opus-5 3/6, GPT-6-astra 2/6, and GPT-5.6-sol and GPT-6-sol 0/6. Figure~\ref{fig:cqa-single-combined} reports question accuracy across individual and combined settings.

\paragraph{Known failure modes.}
\begin{itemize}
  \item Incorrect joins, time windows, or score conventions produce patient-specific calculation errors.
  \item Unimplemented calculations and fallback guesses leave questions incorrect despite successful database access, as illustrated by the failed Opus-5 combined attempt in Appendix~\ref{appendix:trajectories}.
\end{itemize}
\end{revision}

\subsection{\rev{\taskSurvival: cancer time-to-event prediction}}
\label{appendix:bench:survival}
\begin{revision}
\paragraph{Tags.}
\begin{itemize}
  \item \emph{Modality:} Clinical, demographic, laboratory, and genomic tables.
  \item \auditrev{\emph{Workflow:} Research.}
  \item \emph{Output shape:} One predicted survival time per held-out patient.
  \item \emph{Data access:} Public MSK-CHORD cohort files, staged locally before execution.
  \item \emph{Number of Tasks:} 5.
\end{itemize}

\paragraph{Motivation.}
This task category introduces cancer survival prediction from clinical and genomic patient data in \projectname. Estimating cancer survival supports prognostic assessment and research into differences in outcomes across patients.
The agent must build and validate a survival-prediction pipeline from labeled patient tables, then predict outcomes for held-out patients. \projectname adapts Survprompt's \texttt{TTE\_OS} representation and data split~\citep{survprompt} into an autonomous modelling task whose success requires outperforming cohort-matched Cox and random survival forest (RSF) references.

\paragraph{Task setup.}
\begin{itemize}
  \item \emph{Inputs:} \texttt{train.csv}, \texttt{test\_examples.csv}, and \texttt{manifest.json} under \path{/workspace/data/}. Training rows contain features, observed follow-up times, and death/censoring indicators; test rows omit outcomes. The manifest identifies the split and feature columns.
  \item \emph{Instruction summary:}
    \begin{quote}\small\itshape
    Build a survival model from the labeled training patients. For every held-out patient, predict survival time in days from the available features. You may choose and refine any modelling approach; only the final predictions are scored. Treat patient identifiers only as submission keys. You have up to one hour.
    \end{quote}
  \item \emph{Required action:} process missing values and features, fit and validate a model using training data, and generate predictions for all test patients.
  \item \emph{Output:} \path{/workspace/submission/predictions.csv}, with columns \texttt{sample\_id,prediction}; each expected identifier must appear exactly once with a finite, nonnegative survival time in days.
  \item \emph{Limits:} agent timeout 3600~s, verifier timeout 300~s, 16~CPUs, 32~GiB RAM, 32~GiB storage, no GPU, and model API access only.
\end{itemize}

\paragraph{Data source.}
\begin{itemize}
  \item \emph{Corpus:} public MSK-CHORD clinical/genomic cohort tables~\citep{mskchorddata}. Outcomes measure time from first sequencing to death or censoring.
  \item \emph{Split:} exact duplicate rows are removed before selecting fold~0 of Survprompt's five-fold split, with seed~20.
  \item \emph{Feature convention:} supplied progression and summary laboratory fields are treated as a fixed landmark snapshot.
\end{itemize}

\paragraph{Tasks in this Category.}
\begin{itemize}
  \item \emph{Count:} five time-to-event tasks, one each for breast, colorectal, non-small-cell lung, pancreatic, and prostate cancer.
  \item \emph{Cohort sizes:} train/test counts are 4,294/1,074; 4,434/1,109; 6,247/1,562; 2,487/622; and 2,568/643, respectively.
\end{itemize}

\paragraph{Scoring.}
\begin{itemize}
  \item \emph{Reward (Success Criterion):} full valid coverage and strict improvement over both Cox and RSF on all three metrics: higher Harrell's C-index, lower confidence-weighted MAE-PO (cMAE), and lower Graf IPCW integrated Brier score (IBS). Missing, duplicate, unexpected, or malformed predictions prevent a pass.
  \item \emph{Raw Per-trial Metrics:} coverage, C-index, cMAE in years, and IBS (Figure~\ref{fig:survival-tte}).
  \item \emph{Metric conventions:} C-index uses negative predicted time as risk. MAE-PO uses training outcomes as its Kaplan--Meier reference and converts days to years using 365.25 days/year. IBS converts scalar predictions to step survival curves and estimates censoring from training outcomes. Its half-year grid extends to at most ten years, bounded by observed follow-up; this grid does not cap submitted times. The task instructions specify the integration and tied-time conventions.
  \item \emph{References:} Cox uses penalization 0.1; RSF uses 1,000 trees, minimum split size 10, minimum leaf size 15, and seed~20. Both share the agent's cohort, split, features, prediction representation, and metric implementation.
\end{itemize}

\paragraph{Anti-cheat \& leak-proofing.}
\begin{itemize}
  \item \emph{Outcome isolation:} raw source files and held-out outcomes are available only to bootstrap and the verifier. The agent receives labeled training data and unlabeled test features.
  \item \emph{Identifier restriction:} identifiers are opaque join keys; their values, numeric components, ordering, and transformations must not inform prediction, calibration, or model selection.
\end{itemize}

\paragraph{Bootstrap \& environment.}
\begin{itemize}
  \item \emph{Data staging:} users place public source files in the host cache; bootstrap constructs the task split and separates agent-visible tables from verifier-only outcomes. The source files are not redistributed with the suite.
  \item \emph{Tools:} pandas, NumPy, SciPy, scikit-learn, scikit-survival, and lifelines are preinstalled; the verifier runs offline in a separate container.
\end{itemize}

\paragraph{Baseline observations.}
With one attempt per task, GPT-6-astra, GPT-6-sol, and Opus-5.5 pass 5/5; Sonnet-5.5 passes 4/5; GPT-5.6-sol and Opus-5 pass 3/5 (Appendix~\ref{appendix:per-task:survival}).

\paragraph{Known failure modes.}
\begin{itemize}
  \item Validation performance may not generalize to held-out patients: all five failed runs miss the IBS threshold, and Opus-5's pancreatic run also misses cMAE. Two GPT-5.6-sol runs contain small local metric-implementation mistakes, but their unchanged final predictions fail under both local and official metrics.

\end{itemize}
\end{revision}

\section{\rev{Trajectory Analysis of Selected Runs}}
\label{appendix:trajectories}

\begin{revision}
For each category, we describe one successful and one failed attempt from the selected v2 results, using saved trajectories, submissions, and verifier outputs. Success and failure refer to the benchmark's recorded decisions; these examples illustrate observed strategies. Failure attributions identify the weakness observed in each attempt.

\subsection{X-ray Report Correction}
\paragraph{Successful attempt.}
On case~02, Sonnet-5.5 reads the draft and prior reports, inspects the target radiograph, and makes three targeted corrections: stable rather than worsening cardiomegaly, aortic-arch rather than mitral-valve calcification, and absence rather than presence of overt pulmonary edema. It saves the corrected report in the required format. All five saved judge votes find no significant error, so the attempt passes.

\paragraph{Failed attempt.}
On case~01, Sonnet-5.5 also inspects the images and revises the report, but places emphysematous lucency in the lower lobes, whereas the reference report places it in the upper lobes. All five judges flag this location discrepancy as significant. The apparent weakness is anatomical localization and consistency between the image interpretation and report wording. All judge calls completed, but the report-derived reference and LLM judgments have not been independently clinically adjudicated, so a reference-label issue cannot be ruled out.

\subsection{CT Abnormality Classification}
\paragraph{Successful attempt.}
On volume~\texttt{valid\_16\_a\_1}, Opus-5.5 loads the CT and examines axial slices using lung and soft-tissue windows. It identifies lung opacity and consolidation, while rejecting cardiomegaly, pericardial effusion, and lymphadenopathy. All five predictions match the retained reference labels, satisfying the all-correct gate.

\paragraph{Failed attempt.}
On volume~\texttt{valid\_137\_a\_1}, Opus-5.5 inspects multiple planes and measures attenuation in suspicious regions. It correctly predicts five of six labels but rejects hiatal hernia, which is positive in the reference. The apparent weakness is recognition of the relevant anatomy and finding, rather than volume access or rendering. The all-correct gate converts one disagreement into failure; independent clinical review would be needed to distinguish an interpretive miss from a report-derived label issue.

\subsection{\taskMri}
\paragraph{Successful attempt.}
On set~02, Opus-5.5 renders the supplied arrays in three planes, inspects ligament and meniscus regions, and requests additional image readings from internal subagents. It reconciles disagreements and submits ACL and meniscus answers for all three examinations. All six answers match the reference, including the meniscus findings it identifies as uncertain.

\paragraph{Failed attempt.}
On set~03, Opus-5.5 again uses multiplanar inspection and additional readers, revisiting disputed meniscus findings before submission. It predicts no meniscal tear in examination~1, contrary to the reference, while the other five answers are correct. The apparent weakness is resolving an uncertain meniscal finding despite additional readers and image inspection.

\subsection{Pathology Tumor Area Selection}
\paragraph{Successful attempt.}
On slide~0001, Sonnet-5.5 begins with a whole-slide overview, inspects higher-resolution crops, and develops a local mask using color and nuclear-density features. It revises the mask after checking overlays and tumor boundaries, then converts it into tile coordinates. The submission has 162 true-positive, seven false-positive, and one false-negative tiles, giving F1~0.9759 above the required 0.9.

\paragraph{Failed attempt.}
On slide~0002, Sonnet-5.5 similarly uses image inspection and a color-based mask to select tumor tiles. Its final set includes all 141 reference-positive tiles but also 47 false positives. F1 falls to 0.8571. The observed weakness is precise tumor-boundary delineation: the mask overselects tissue relative to the reference despite complete tumor-tile recall. The saved evidence supports a localization-precision shortfall.

\subsection{Pathology Tumor Slide Selection}
\paragraph{Successful attempt.}
On set~05, Sonnet-5.5 examines thumbnails and higher-resolution tissue crops and records a separate morphology-based judgment for each slide. It selects slides~1, 3, 6, and~8 as tumor-bearing. This exactly matches the reference set, so the attempt passes.

\paragraph{Failed attempt.}
On set~08, Sonnet-5.5 also examines multiple magnifications but interprets slide~6 as non-tumor tissue. It selects slides~3, 4, and~8, omitting reference-positive slide~6. The partial F1 is 0.8571, but the exact-set criterion fails. The apparent weakness is distinguishing tumor from non-tumor morphology in the inspected tissue.

\subsection{EHR Data Quality Auditing}
\paragraph{Successful attempt.}
On the combined standard task, Opus-5.5 inspects table schemas and checks value ranges, units, duplicated observations, prescription dose/strength consistency, and demographic and cross-table contradictions. It verifies the submitted row identifiers against the source tables and flags 156 rows. These recover all 69 required error clusters with row precision~0.7949, satisfying both scoring conditions.

\paragraph{Failed attempt.}
On the same task, Sonnet-5.5 performs substantive checks for conflicting records, implausible measurements, unit errors, and demographic inconsistencies, submitting 123 rows. Its precision is 0.8049, but it recovers only 58 of 69 required clusters: ten prescription decimal-shift clusters and one prescription-related demographic conflict are missed. The observed weakness is systematic audit coverage, particularly prescription checks and reconciliation across related fields.

\subsection{EHR Compositional QA}
\paragraph{Successful attempt.}
On the combined task, Opus-5.5 assigns groups of clinical-score calculations to four internal subagents and implements the vasopressor-dose calculations itself. It uses supplied training examples to resolve conventions, checks intermediate results, and merges the answers into one submission. All 51 answers are correct.

\paragraph{Failed attempt.}
Opus-5 completes the same combined task without launching subagents and stops normally after 13.79 minutes of a 60-minute allowance. It computes 27 answers, of which 25 are correct, and fills 24 unimplemented calculations with the most frequent corresponding training answers, only two of which are correct. The resulting 27/51 score reflects incomplete implementation and fallback guessing, rather than a recorded timeout. The observed weakness is planning and carrying a multi-part implementation through to completion.

\subsection{EHR Event Modelling}
\paragraph{Successful attempt.}
For hypertension, Sonnet-5.5 constructs features from events preceding each prediction time, including diagnosis and medication counts, demographic variables, and laboratory and vital-sign summaries over several look-back windows. It compares models using patient-grouped validation and submits a blend of LightGBM and regularized logistic regression for all 755 test rows. AUROC is 0.7463, exceeding the Count+GBM reference of 0.6725.

\paragraph{Failed attempt.}
For pancreatic cancer, Sonnet-5.5 engineers temporal code-count and laboratory features and reconstructs additional labeled examples from training/validation patients while excluding test patients. After grouped validation, it submits a rank-averaged LightGBM/logistic-regression ensemble for all 1,246 test rows. The final pipeline corrects visit-end capping, but its test AUROC of 0.8977 remains below the 0.9204 reference. The observed shortfall is predictive discrimination on held-out patients: the chosen features and model do not rank outcomes as well as the reference.

\subsection{Cancer Survival Prediction}
\paragraph{Successful attempt.}
For breast-cancer time-to-event prediction, GPT-6-astra compares hazard and boosted Cox/AFT models, combines their predictions, and calibrates scalar survival times using repeated validation folds. It submits finite, nonnegative predictions for all 1,074 test patients. C-index is 0.8091, censoring-adjusted MAE is 2.0543 years, and IBS is 0.2000; all three improve on both reference pipelines, whose stricter thresholds are 0.7786, 2.2509 years, and 0.2390, respectively.

\paragraph{Failed attempt.}
For lung-cancer time-to-event prediction, Opus-5 fits an elastic-net Cox and boosted-Cox ensemble, then calibrates risk scores into survival times using out-of-fold predictions. Its validation IBS is 0.2568, but the submitted predictions yield test IBS~0.2681, above the required 0.2614, despite passing the C-index and error gates. Reconstructing the agent's metric helper on the saved predictions matches the official scores, supporting a held-out performance shortfall rather than a metric-definition error in this attempt. The observed weakness is generalization of the risk-to-time calibration: acceptable discrimination does not translate into sufficiently accurate event-timing predictions under IBS.
\end{revision}

\end{document}